%% file: main.tex
\documentclass[11pt]{article}

\usepackage[final]{acl}

\usepackage{times}
\usepackage{latexsym}
\usepackage[T1]{fontenc}
\usepackage[utf8]{inputenc}
\usepackage{microtype}
\usepackage{inconsolata}
\usepackage{graphicx}
\usepackage{subcaption}
\usepackage{amsmath}
\usepackage{booktabs}
\usepackage{multirow}
\usepackage[table]{xcolor}
\usepackage{url}
\usepackage{hyperref}
\usepackage{enumitem}
\usepackage{pifont}
\usepackage{afterpage}
\usepackage{flafter}  
\newcommand{\cmark}{\textcolor{green!60!black}{\ding{51}}}
\newcommand{\xmark}{\textcolor{red}{\ding{55}}}

\title{MemUse: Moving Memory Evaluation from Direct QA\\
to Natural Integration in Long-Term Human-AI Conversation}

\author{
  Ryuichi Sumida \quad Koji Inoue \quad Tatsuya Kawahara \\
  Graduate School of Informatics, Kyoto University \\
  \texttt{sumida.ryuichi.65m@st.kyoto-u.ac.jp} \\
  \texttt{inoue@sap.ist.i.kyoto-u.ac.jp}, \texttt{kawahara@i.kyoto-u.ac.jp}
}

\begin{document}
\maketitle

\input{sections/00_abstract}
\input{sections/01_introduction}
\input{sections/02_related_work}
\input{sections/03_study_design}
\input{sections/04_puzzle}      
\input{sections/05_memuse}      
\input{sections/06_findings}    
\input{sections/07_discussion}  

\section*{Limitations}
\input{sections/08_limitations}

\section*{Ethics Statement}
This study was approved by our institutional review board.
All participants provided informed consent prior to participation and were compensated for their time; consent explicitly authorized public release of their conversations and satisfaction ratings for research use.
Participants were instructed to self-curate their disclosures, sharing only content they were comfortable releasing publicly.
A one-month withdrawal window at the start of the 4-month deployment allowed participants to revoke consent; one participant did so and their data is excluded from the reported sample.
Prior to release, the full deployment corpus was screened with GPT-5.4 for residual direct identifiers---personal names, locations, workplaces, and contact information referring to participants or to third parties---and flagged spans were reviewed by the research team and redacted.
Raw conversation data was stored on encrypted institutional servers with access restricted to the research team during the study; human annotators who verified memory events accessed only de-identified transcripts.
We release the full deployment corpus (1{,}872 sessions across 40 users and 7 memory conditions, with per-session satisfaction ratings), the \textsc{MemUse} benchmark (72~de-identified instances with ground-truth facts and questions), and all evaluation and analysis code at \url{https://huggingface.co/datasets/RuiSumida/memuse} and \url{https://github.com/ryuichi-sumida/memuse} (release notes in Appendix~\ref{app:data_release}).

\section*{Acknowledgments}
This work was supported by JST BOOST Grant Number JPMJBS2407, JST Moonshot R\&D JPMJPS2011, and NEXUS JPMJNX25C1.

\bibliography{references}

\appendix
\input{sections/09_appendix}

\end{document}

%% file: sections/00_abstract.tex
\begin{abstract}
Memory systems for conversational LLMs are conventionally evaluated by direct, fact-seeking questions about prior dialogue (\textit{Direct~QA}): \textit{can the model recall fact X from a prior conversation?}
We tested whether higher Direct~QA accuracy correlates with higher user satisfaction in a 4-month deployment (40~users, 1{,}872~sessions, 7~memory conditions).
Existing-benchmark Direct~QA varies from 19.7\% to 70.1\% across the 7 conditions, but satisfaction does not change.
We hypothesize that existing benchmarks and user satisfaction are tracking different capabilities: benchmarks measure \textit{elicited retrieval} (recall when asked), while conversation requires \textit{natural integration} (detecting relevance and naturally weaving prior context into a response).
To examine this, we introduce \textsc{MemUse}, a set of real user-cued memory moments drawn from the deployment, scored by an integration-aware judgment of the natural conversational response.
Holding the model and context fixed, the same system that scores 78.8\% on Direct~QA references only 7.9\% of those facts in conversation---a 71-point gap.
Within these moments, Natural Integration is associated with satisfaction, whereas Direct~QA is not.
We release the deployment corpus and \textsc{MemUse} together with all judgments and scoring prompts at \url{https://github.com/ryuichi-sumida/memuse}.
\end{abstract}

%% file: sections/01_introduction.tex
\section{Introduction}

As LLMs are deployed in long-term conversational settings---personal assistants, therapy bots, diary companions---their memory of past interactions is assumed critical for user satisfaction~\cite{skjuve2021chatbotcompanion,brandtzaeg2022myai}.
Recent benchmarks like LoCoMo~\cite{maharana2024evaluating}, LUFY~\cite{sumida2025lufy}, and RealTalk~\cite{lee2025realtalk} evaluate memory through direct, fact-seeking questions about prior dialogue (we refer to this evaluation format as \textit{Direct~QA} throughout): \textit{can the model recall fact X from a prior conversation?}
Memory systems have accordingly focused on expanding capacity through longer contexts~\cite{chen2023extending} and retrieval augmentation~\cite{lewis2020retrieval}.
All share an untested structural assumption: that recall-when-asked (\textit{elicited retrieval}) reflects whether a model will use prior knowledge in conversation (\textit{natural integration}).

\begin{table*}[!t]
\centering
\footnotesize
\setlength{\tabcolsep}{4pt}
\begin{tabular}{@{}llrrrlrcc@{}}
\toprule
\textbf{Benchmark} & \textbf{Authoring} & \textbf{Users} & \textbf{Sess./u.} & \textbf{Turns/u.} & \textbf{Date span} & \textbf{Mem.\ \%}$^\dagger$ & \textbf{Nat.\ cued} & \textbf{Sat.} \\
\midrule
\multicolumn{9}{@{}l}{\textit{Synthetic / LLM-simulated dialogues}} \\
MemoryBank~\cite{zhong2024memorybank} & authors & 15 & 10 & 37.7 & 10 days & 17.7\% & \xmark & \xmark \\
LoCoMo~\cite{maharana2024evaluating} & authors & 20 & 27.2 & 294.1 & $\sim$2 mo & 24.0\% & \xmark & \xmark \\
LongMemEval~\cite{wu2025longmemeval} & experts & 500 & 47.7 & 244.8 & --- & 0.4\% & \xmark & \xmark \\
\midrule
\multicolumn{9}{@{}l}{\textit{Real human conversation}} \\
LUFY~\cite{sumida2025lufy} & users & 17 & 4 & 123.1 & $\sim$9 days & 23.8\% & \xmark & \xmark \\
RealTalk~\cite{lee2025realtalk} & crowd & 20 & 21.9 & 193.7 & $\sim$3 weeks & 14.6\% & \xmark & \xmark \\
\textsc{MemUse} (Ours) & detected & 40 & 46.8 & 269.9 & $\sim$4 months & 1.4\% & \cmark & \cmark \\
\bottomrule
\end{tabular}
\caption{Long-term memory benchmarks. \textsc{MemUse} is the only one combining multi-month real human-AI deployment, evaluation items detected from the conversation rather than authored as QA, and per-session satisfaction ratings. Column abbreviations: \textbf{Sess./u.}~=~sessions per user; \textbf{Turns/u.}~=~user turns per user; \textbf{Mem.\ \%}$^\dagger$~=~\% of user turns paired with an evaluation item; \textbf{Nat.\ cued}~=~evaluation items naturally cued by users (vs.\ externally authored); \textbf{Sat.}~=~per-session user satisfaction rating available. LoCoMo and RealTalk pair the same script across two speakers, so their 20 users represent 10 distinct dialogues.}
\label{tab:dataset_comparison}
\end{table*}

But does benchmark performance on Direct~QA actually translate to user satisfaction in real-world settings?
We conducted a 4-month longitudinal study where 40~users interacted daily with an AI diary companion across 7~memory conditions, ranging from summary-only to full long-context and RAG systems.
At the population level, higher Direct~QA accuracy does not raise satisfaction (\S\ref{sec:result1_null}): condition-level mean differences against Summary-only are all below 0.06 within-user~SD.

The null is the puzzle, not the conclusion. We hypothesize that existing benchmarks and user-experience measures are simply tracking different capabilities: benchmarks measure \textit{elicited retrieval}, while conversation requires \textit{natural integration}.
To examine this, we introduce \textsc{MemUse} (\S\ref{sec:benchmark}), a benchmark of 72~real moments where users explicitly referenced or tested the system's memory, scored by a single \textit{Natural Integration} judgment of whether the system's natural response demonstrates memory of the referenced topic.

\paragraph{Contributions.}
We make three contributions:
\begin{itemize}
  \item A randomized 4-month deployment (40 users, 1{,}872 sessions, 7 memory conditions) in which Direct~QA accuracy varies from 19.7\% to 70.1\% across conditions, yet user satisfaction does not change (\S\ref{sec:result1_null}).
  \item \textsc{MemUse}, a benchmark of 72~real user-cued memory moments, scored by whether the model's conversational response actually draws on the referenced memory---a measure that, unlike Direct~QA, is associated with satisfaction (\S\ref{sec:benchmark}, \S\ref{sec:integration_predicts}).
  \item Evidence that retrieval and integration are dissociable: the same system answers 78.8\% of \textsc{MemUse} items correctly under Direct~QA, but references those facts in conversation only 7.9\% of the time---a 71-point gap (\S\ref{sec:gap}).
\end{itemize}

%% file: sections/02_related_work.tex
\section{Related Work}

\paragraph{Memory Benchmarks.}
Existing memory benchmarks---LoCoMo~\cite{maharana2024evaluating}, RealTalk~\cite{lee2025realtalk}, LUFY~\cite{sumida2025lufy}, and LongMemEval~\cite{wu2025longmemeval}---rely on synthetic or externally authored questions in a Direct~QA format and none validate whether their scores predict user experience (Table~\ref{tab:dataset_comparison}).

\paragraph{Memory-Augmented Systems.}
Personalized dialogue systems combine summarization~\cite{lu2023memochat}, retrieval~\cite{lewis2020retrieval}, and long-context prompting~\cite{chen2023extending} with explicit memory management~\cite{packer2023memgpt,chhikara2025mem0,ong2025theanine}; selective forgetting can preserve performance at lower cost~\cite{sumida2025lufy}, which we use to instantiate our LC/RAG capacity range.
Long-context prompting carries integration risk---models often ignore or degrade on relevant evidence~\cite{liu2024lostmiddle}---a phenomenon our retrieval--integration gap (\S\ref{sec:gap}) extends to natural conversation.
Evaluation across these systems still centers on recall accuracy; we instead test these strategies against longitudinal user satisfaction.

\paragraph{Dialogue Evaluation Beyond Accuracy.}
Disconnects between automatic metrics and human judgment are well documented in dialogue~\cite{deriu2021survey}, with prior work on satisfaction prediction~\cite{bodigutla2020joint,santana2025speechtojoy} and long-term human-chatbot relationships~\cite{skjuve2021chatbotcompanion} highlighting user experience beyond benchmark accuracy.

%% file: sections/03_study_design.tex
\section{Experimental Setup}
\label{sec:study_design}

Forty proficient English speakers from diverse geographic backgrounds (approximately half based in East Asia, remainder from North America, Europe, the Middle East, and other regions; 37F, 3M; ages 20s--60s) interacted daily with ``Luke,'' an AI diary companion built on GPT-4.1-mini, from November 2025 to February 2026.
Users accessed the system through a web interface (compatible with both mobile and desktop) where they could review summaries of prior conversations before starting each session (Appendix~\ref{app:interface}).
Participants were required to interact for at least 1~hour and write around 2{,}000 English words per month; content was unconstrained---users wrote about daily life, personal reflections, and any topic of their choosing.
The study yielded 1{,}872~sessions (21{,}575~turns); users rated each session on a 1--7 satisfaction scale.

Each session was randomly assigned one of 7~memory conditions: \textsc{Summary}, \textsc{LC}-10/50/100\%, or \textsc{RAG}-10/50/100\%.
\textbf{Summary-only} supplies only a summary of prior conversations; all other conditions add capacity on top of this same summary.
\textbf{LC-$k$\%} prepends the top $k$\% most important prior turns to the conversation history.
\textbf{RAG-$k$\%} retrieves the top-10 turns from a vector index over that same top-$k$\% pool.
Importance is scored by a RoBERTa model finetuned on human annotations from the LUFY dataset~\cite{sumida2025lufy}.
On a held-out evaluation (trained on two annotator groups, tested on a third), the model exceeds the human annotator average on both important-utterance identification (56.3\% vs.\ 41.4\%) and not-important identification (93.1\% vs.\ 91.3\%), ranking 7th among 18~raters (17~humans + the model).
Conditions were randomly rotated to ensure balanced exposure (Appendix~\ref{app:balance}); engagement remained stable across the 4-month study (Appendix~\ref{app:engagement_patterns}).
All seven conditions share the same persona and per-session framing; verbatim generation prompts are in Appendix~\ref{app:prompts_verbatim}.

For satisfaction analysis, we applied a predefined filtering pipeline (Appendix~\ref{app:data_flow}):
from 1{,}872~sessions, we excluded 11~users---9 with near-constant ratings (SD~$\leq 0.5$ on the 1--7 scale) plus 2 flagged for content quality (template/AI-paste content)---yielding 1{,}270~sessions from 29~users.
We analyze within-user $z$-scored satisfaction.

%% file: sections/04_puzzle.tex
\section{Capacity Improves Direct~QA but Not Satisfaction}
\label{sec:result1_null}

\begin{table}[!t]
\centering
\small
\setlength{\tabcolsep}{4pt}
\begin{tabular}{lcccc}
\toprule
\textbf{Cond.} & \textbf{Tok.\,Gr.} & \textbf{vs Sum.} & \textbf{Mean Lat.} & \textbf{Mean TTFT} \\
\midrule
Summary  & 1.0$\times$ &  1.0$\times$ & 1.93 & 1.04 \\
LC-10    & 2.0$\times$ &  2.9$\times$ & 2.00 & 1.12 \\
LC-50    & 3.5$\times$ & 11.9$\times$ & 2.52 & 1.41 \\
LC-100   & 5.8$\times$ & 22.0$\times$ & 3.27 & 1.79 \\
RAG-10   & 1.8$\times$ &  2.3$\times$ & 2.03 & 1.05 \\
RAG-50   & 1.0$\times$ &  2.5$\times$ & 2.03 & 1.09 \\
RAG-100  & 1.4$\times$ &  2.1$\times$ & 2.13 & 1.12 \\
\bottomrule
\end{tabular}
\caption{Per-condition cost and latency on the 29-user / 1{,}270-session subset. \textbf{Tok.\,Gr.}: within-condition input-token growth from month~1 to month~4. \textbf{vs Sum.}: mean session input tokens relative to the Summary baseline. \textbf{Mean Lat.}: per-session mean total response latency (s). \textbf{Mean TTFT}: per-session mean time-to-first-token (s).}
\label{tab:cost}
\end{table}

Our research question is simple: does higher Direct~QA performance---obtained by raising memory capacity beyond a summary baseline---translate into higher user satisfaction? Under the prevailing assumption that benchmark scores reflect user value, LC-100\% should rank first and Summary-only last; we test that prediction directly.

\paragraph{Existing-benchmark accuracy rises sharply with capacity.}
Across the 7~conditions, average existing-benchmark QA accuracy climbs from 19.7\% (Summary) to 70.1\% (LC-100\%) (full per-condition numbers in Appendix~Table~\ref{tab:main_combined}, per-benchmark breakdown in Appendix~\ref{app:qa_breakdown}).

\paragraph{User satisfaction does not follow.}
User ratings do not change with capacity. Every condition's mean satisfaction differs from Summary-only by less than 0.06 within-user~SD (Appendix~Table~\ref{tab:main_combined}).
Numerically, LC-100\% is not better than Summary; the directional ranking has Summary slightly ahead ($+0.02$ vs.\ $-0.04$~SD).
Because every condition includes the same summary, this is a null about the marginal value of LC/RAG capacity on top of a summary, not about memory in general.
This flat curve is not just rating noise. In a session-level linear mixed-effects model, we entered three engagement features---response length, response specificity (the proper-noun rate in the system's reply, as a proxy for whether responses name concrete user details), and cross-session continuity (how much the user's current turn references prior-session topics)---alongside memory condition.
Each engagement feature independently predicts higher satisfaction, while memory condition does not (full models in Appendix~\ref{app:engagement}; robustness in Appendix~\ref{app:null_robustness}).
The null is also not latency-mediated: on the same 29-user subset, neither mean, max, nor min total response latency---nor their TTFT counterparts---correlates with within-user $z$-scored rating ($|\rho|<0.06$, all $p\geq.07$; per-condition latencies in Table~\ref{tab:cost}, full breakdown in Appendix~\ref{app:cost}).

%% file: sections/05_memuse.tex
\section{Detecting and Scoring Memory Moments}
\label{sec:benchmark}

\begin{table*}[!t]
\centering
\small
\setlength{\tabcolsep}{4pt}
\renewcommand{\arraystretch}{1.18}
\definecolor{badcell}{HTML}{FCE4E4}
\definecolor{warncell}{HTML}{FFF4D6}
\definecolor{goodcell}{HTML}{E4F4E4}
\definecolor{ctxcell}{HTML}{EEF1F5}
\definecolor{idealcell}{HTML}{DEE8F5}
\begin{tabular}{@{}p{0.19\textwidth}p{0.77\textwidth}@{}}
\toprule
\multicolumn{2}{@{}p{0.97\textwidth}@{}}{\cellcolor{ctxcell}\textbf{Source sessions (prior context):}\quad In sessions~2--3, the system (``Luke'') recommended Bruno Mars (upbeat) and Ryuichi Sakamoto, Hikaru Utada, Yumi Matsutoya, and Haruka Nakamura (Japanese relaxing music).} \\
\multicolumn{2}{@{}p{0.97\textwidth}@{}}{\cellcolor{ctxcell}\textbf{Trigger (user turn, current session):}\quad \textit{``I'm thinking I'll try to relax by listening to the music Luke recommended the other day.''}} \\
\midrule
\multicolumn{2}{@{}l@{}}{\textbf{(i) Natural Integration} \textit{(primary metric: does the system's natural response demonstrate memory of the referenced topic?)}} \\
LC-100\% response & \cellcolor{warncell}\textit{``That sounds like a perfect way to unwind. I hope the music brings you some calm and comfort while your little one rests. Enjoy this peaceful moment---you deserve it!''} \hfill\xmark~\textsc{\textcolor{orange!85!black}{generic response}} \\
Ideal response (authored) & \cellcolor{idealcell}\textit{``Glad you're going to try the music I recommended! Ryuichi Sakamoto's \emph{Merry Christmas Mr. Lawrence} is my favorite---the solo piano eases you down.''} \hfill\cmark~\textsc{\textcolor{green!40!black}{integrates context}} \\
\midrule
\multicolumn{2}{@{}l@{}}{\textbf{(ii) Direct~QA} \textit{(same model, same reconstructed context, each fact-seeking question asked literally; excerpt of 4)}} \\
Q1 & \textit{Q: ``What upbeat artist did the system recommend?''} \newline \cellcolor{goodcell}\textit{A: ``Bruno Mars.''} \hfill\cmark \\
Q2 & \textit{Q: ``What Japanese artists did the system recommend for relaxing music?''} \newline \cellcolor{goodcell}\textit{A: ``Ryuichi Sakamoto, Hikaru Utada, Yumi Matsutoya, and Haruka Nakamura.''} \hfill\cmark \\
\midrule
\multicolumn{2}{@{}l@{}}{\textbf{(iii) Reference} \textit{(do those facts actually appear in the natural response above?)}} \\
Reference & \cellcolor{badcell}0~/~4 facts referenced in the LC-100\% response. \hfill\xmark \\
\bottomrule
\end{tabular}
\caption{One real \textsc{MemUse} instance scored under all three metrics (GPT-4.1-mini, LC-100\% condition, same reconstructed context throughout). The same model that fails (i)~Natural Integration and (iii)~Reference answers the underlying questions correctly under (ii)~Direct~QA---the retrieval--integration gap of \S\ref{sec:gap}.}
\label{tab:memuse_instance}
\end{table*}

The deployment puzzle (\S\ref{sec:result1_null}) motivates a closer look at the rare moments where memory is actually at stake: if memory matters to users, the effect should appear when users explicitly cue or encounter memory use---not averaged across sessions where memory was never invoked.
We audit the 1{,}872-session corpus for these moments and package the user-cued subset as the \textsc{MemUse} benchmark.

\subsection{Detecting Memory Moments}
\label{sec:detection}

\paragraph{Detection.}
We extract memory-relevant moments using GPT-5.4 (temperature=0) on each of 1{,}872~session transcripts, detecting four event types through explicit linguistic signals:
\textbf{user memory probes} (``Do you remember X?''),
\textbf{user re-provisions} (``As I mentioned before, X''),
\textbf{system proactive recalls} (``How was your trip to the flower shop?''), and
\textbf{user memory reactions} (``You remembered!'').
Detected events were independently verified by human annotators, who reviewed each candidate alongside (i)~the trigger utterance, (ii)~the surrounding conversation context (the current session plus summaries of the three preceding sessions), and (iii)~the detector-supplied memory rationale (the prior-session content the detector identified as the referent and its justification for flagging the moment).
This yielded 147 true positives---11~probes, 62~re-provisions, 66~proactive recalls, 8~reactions---at 95.5\% precision (full procedure and per-type breakdown in Appendix~\ref{app:detection}).
\afterpage{%
\begin{figure*}[!t]
\centering
\includegraphics[width=\textwidth]{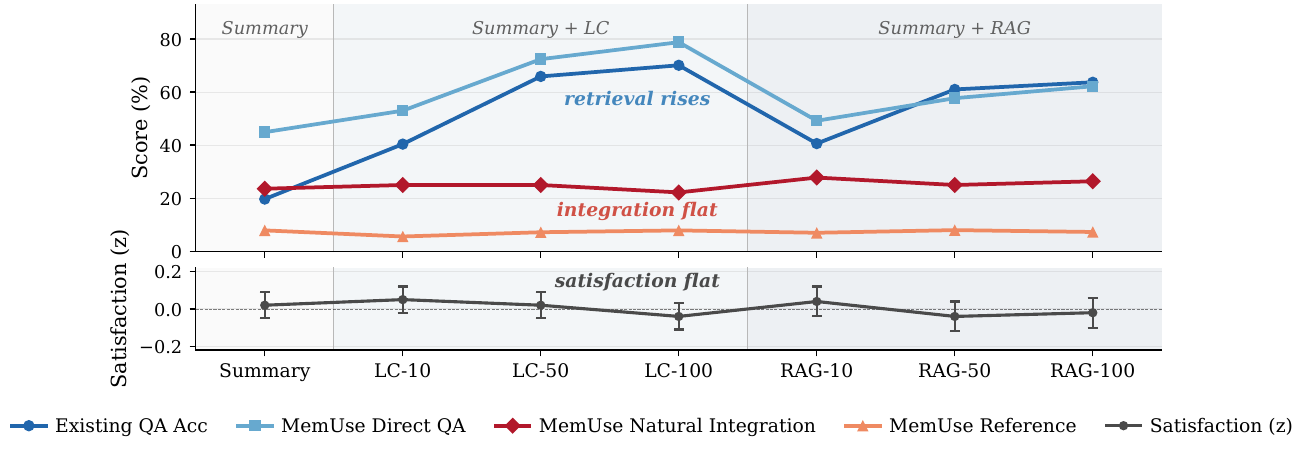}
\caption{Across 7 randomized memory conditions (40 users, 1{,}872 sessions; \textsc{LC} prepends the top-$k$\% most important prior turns, \textsc{RAG} retrieves from the same top-$k$\% pool; full definitions in \S\ref{sec:study_design}), the two retrieval metrics (Existing QA Acc; \textsc{MemUse} Direct~QA) rise sharply with capacity, while the integration metrics (\textsc{MemUse} Natural Integration; \textsc{MemUse} Reference) stay flat. Satisfaction (lower panel; 1{,}270 rating-valid sessions, 29 users) is shown for comparison. Numerical values in Table~\ref{tab:main_combined} (Appendix~\ref{app:result1_detail}).}
\label{fig:story}
\end{figure*}%
}

\paragraph{Memory moments are rare.}
Across the 1{,}872 sessions (10{,}795~user turns total; mean~5.8 per session), the 147 verified events arise in only $\sim$3.5\% of sessions and 1.4\% of user turns---one detected memory moment per~$\sim$73 user turns.
This 1.4\% rate is the natural frequency at which users invoke memory in deployment; existing benchmarks instead pair externally authored QA with much higher fractions of user turns (15--24\% across MemoryBank, LoCoMo, LUFY, and RealTalk; Table~\ref{tab:dataset_comparison}).

\paragraph{Reactive vs.\ proactive split.}
The four event types separate cleanly by who initiates the memory use.
\textbf{Reactive (user-cued)} moments are user-initiated invitations to use prior context (analyzed in \S\ref{sec:integration_predicts}).
\textbf{Proactive (system-initiated)} recalls are turns where the system refers to prior-session content unprompted (analyzed in \S\ref{sec:proactive}).
We package only the reactive cases as a re-runnable benchmark, for two reasons.
First, each reactive moment has a clear per-instance target---the user has explicitly elicited a specific piece of prior context, so any candidate memory system can be re-tested on the same trigger and scored against the same target.
Proactive recalls have no such fixed target---there is no canonical ``correct'' next turn---so they cannot be re-played across alternative memory systems in a comparable way.
Second, only reactive integration is related to session-level satisfaction in our deployment; neither proactive recall accuracy nor appropriateness is (\S\ref{sec:integration_predicts}, \S\ref{sec:proactive}). Improvements on this benchmark therefore have a plausible path to user-visible benefit.

\subsection{The \textsc{MemUse} Benchmark}
\label{sec:memuse_benchmark}

\textsc{MemUse} is the reactive benchmark packaged for re-evaluation: 62~re-provisions $+$ 10~probes $=$ 72~instances.
For each instance we extracted ground-truth facts from the source sessions and decomposed them into 3--5 fact-seeking questions (316~total, avg.~4.4 per instance), phrased in the same format used by existing QA benchmarks.
Given an instance and a candidate memory system, the benchmark scores three metrics on the same reconstructed context (Table~\ref{tab:memuse_instance}):
(i)~\textbf{Natural Integration} (the primary metric)---a per-instance binary judgment of whether the system's natural conversational response to the trigger demonstrates memory of the referenced topic, produced by GPT-5.4-nano at temperature~0 following~\citet{zheng2023judging};
(ii)~\textbf{Direct~QA}---the same Direct~QA format used by existing benchmarks (\S\ref{sec:result1_null}). Each fact-seeking question is asked literally with the same reconstructed context, making this the apples-to-apples analogue of existing benchmarks on real memory moments;
(iii)~\textbf{Reference}---a per-question check that the facts elicited by Direct~QA are actually referenced in the natural conversational response.
The metrics~(ii) and~(iii) operate on the same fact-seeking questions and together diagnose the retrieval--integration gap (\S\ref{sec:gap}); Natural Integration is the primary metric because it captures the user-side question---\textit{did the response acknowledge what I just referenced?}---rather than a fact-checker's question (verbatim prompts in Appendix~\ref{app:prompts}).

\paragraph{Judge validation.}
Because LLM judges can be biased~\cite{panickssery2024evaluators}, we validate ours against humans.
On a stratified 56-item subset, human--human agreement on Natural Integration was substantial (Cohen's $\kappa = 0.57$, 95\% bootstrap CI~$[0.34, 0.78]$), and the LLM judge matched the humans on positive rate (LLM 51.8\% vs.\ humans 51.8\%~/~55.4\%).
Per-question judgments showed strong three-rater agreement on the full 316-item set (Fleiss' $\kappa = 0.65$; pairwise human--human Cohen's $\kappa = 0.70$; strictness analyses and per-rater breakdown in Appendix~\ref{app:judge}).

%% file: sections/06_findings.tex
\section{Results}
\label{sec:findings}

Across the 7 memory conditions, retrieval rises sharply with capacity while integration stays flat and satisfaction tracks neither (Figure~\ref{fig:story}). The deployment puzzle (\S\ref{sec:result1_null}) and the memory-moment audit (\S\ref{sec:benchmark}) together motivate three questions about this gap. \textbf{Q1:} at the rare moments where memory is at stake, is \textit{any} measurable system behavior associated with satisfaction? \textbf{Q2:} is that integration ability correlated with what existing benchmarks measure? \textbf{Q3:} does system-initiated memory use deliver the same satisfaction benefit as user-cued integration? We answer them in turn.
All inferential models are linear mixed-effects models (LMMs) with user random intercepts; rank correlations are Spearman's $\rho$, and effect-size confidence intervals are user-clustered bootstrap. Full specifications in Appendix~\ref{app:stats}.

\subsection{Q1: User-Cued Integration Is Associated With Satisfaction}
\label{sec:integration_predicts}
\label{sec:within_instance}
\label{sec:integration_effect}

The puzzle (\S\ref{sec:result1_null}) is that capacity-driven retrieval gains do not change ratings on average.
Restricting attention to the rare sessions where users explicitly cued memory tells a different story (real excerpts in Table~\ref{tab:asymmetry_examples}).
On the 48 reactive memory-moment sessions (after filtering), we evaluate GPT-4.1-mini on two metrics---\textsc{MemUse} Direct~QA and Natural Integration---using the same instances and the same context.

\paragraph{Only Natural Integration is associated with satisfaction.}
Direct~QA accuracy is uncorrelated with session-level satisfaction ($\rho = +0.03$); Natural Integration is associated ($\rho = +0.29$, $p = .046$), with successful integration carrying $+0.56$ within-user~SD higher satisfaction (user-clustered bootstrap 95\% CI $[+0.12, +0.98]$; robust to aggregation rule and engagement covariates, Appendix~\ref{app:nat_aggregation},~\ref{app:amplifier}).
The two metrics also disagree on condition rankings: LC-100\% ranks highest under Direct~QA (78.8\%) and lowest under Natural Integration (22.2\%; visible as the LC-100\% bars in Figure~\ref{fig:story}, per-condition numbers in Appendix~Table~\ref{tab:main_combined}).
A within-user regression with user-level clustering, entering both predictors jointly, finds that Direct~QA adds no predictive value once Natural Integration is included (full specification in Appendix~\ref{app:stats}).

\paragraph{The user-side failure mode: re-provisioning burden.}
Across all 1{,}270~rating-valid sessions, user word count interacts negatively with cross-session continuity ($\beta = -0.086$, $p < .001$; Appendix~\ref{app:reprov}): in high-continuity sessions (those where users reference topics from prior sessions), longer user utterances predict \textit{lower} satisfaction.
This corroborates the integration result with complementary observational evidence: when memory \textit{is} integrated successfully, satisfaction is higher (above); when the user is forced to re-supply context the system has lost, satisfaction is lower.

\subsection{Q2: Why Direct~QA Misses It: Retrieval and Integration Are Dissociable}
\label{sec:gap}

\begin{figure}[!t]
\centering
\includegraphics[width=\columnwidth]{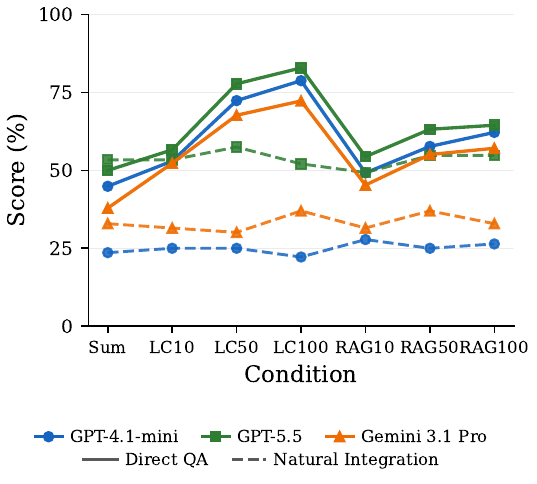}
\caption{The retrieval--integration dissociation replicates across stronger models. \textbf{Direct~QA}: per-question accuracy when each \textsc{MemUse} question is asked literally with the reconstructed context. \textbf{Natural Integration}: reader-level binary judgment of whether the conversational response demonstrates memory of the topic. Numerical values in Table~\ref{tab:cross_model}.}
\label{fig:cross_model}
\end{figure}

The integration-aware result above raises a sharper question: are Direct~QA accuracy and Natural Integration two views of the same underlying ability, or are they dissociable?
We test the stronger claim that they are dissociable---the same model, with the same context, retrieves and integrates the same facts with no detectable correlation.

\paragraph{A concrete example.}
Table~\ref{tab:memuse_instance} shows the gap on a single \textsc{MemUse} instance: under LC-100\%, GPT-4.1-mini's natural response is generic and ignores the user's music-recommendation cue, while the same model with the same context, asked the corresponding fact-seeking question directly, names four of the recommended artists.

\paragraph{Available $\neq$ conversationally integrated.}
\label{sec:gap_available}
In the LC-100\% condition, by design, every prior turn is included in context, and the same model achieves \textbf{existing-benchmark QA accuracy of 70.1\%} on LoCoMo/LUFY/RealTalk with this context, confirming the information is retrievable when explicitly queried.
Yet in natural conversation, GPT-4.1-mini demonstrates \textbf{\textsc{MemUse} Natural Integration of only 22.2\%} of real memory moments (LC-100\% column of Figure~\ref{fig:story}; per-condition numbers in Appendix~Table~\ref{tab:main_combined}).
The same model recovers the fact when explicitly asked, yet does not reference it when the user implicitly cues it.
Classifying all 72 LC-100\% responses by failure mode reveals the dominant patterns: 38.9\% are \textit{generic} (empathetic responses ignoring context), 23.6\% \textit{hallucinate} recall with fabricated details~\cite{huang2025hallucination}, 26.4\% show \textit{partial} (gist-only) recall, 2.8\% are \textit{honest admissions} of not remembering, and only 8.3\% \textit{fully integrate} prior details into their response (Appendix~\ref{app:failures}).

\paragraph{A same-model QA ablation.}
\label{sec:gap_ablation}
\label{sec:gap_perinstance}
To quantify whether the system references the same facts in natural conversation that it can answer when asked directly, we now compare \textsc{MemUse} Direct~QA against \textsc{MemUse} Reference on the same 316 items: GPT-4.1-mini answers each question in Direct~QA format using the identical reconstructed contexts.
The gap is stark: \textbf{\textsc{MemUse} Direct~QA (78.8\%) vs.\ \textsc{MemUse} Reference (7.9\%)---a 71-point gap with the same model and context} on LC-100\% (the gap between the Direct~QA and Reference bars at LC-100\% in Figure~\ref{fig:story}; per-condition results in Appendix~Table~\ref{tab:main_combined}). The two metrics also pull condition rankings apart: condition differentiation invisible in natural responses (22--28\% Natural Integration) emerges clearly in Direct~QA (45--79\%).
Per-instance, the Spearman correlation between Direct~QA correctness and whether the \textit{same fact} is referenced in the natural response is $\rho = -0.009$ (48 deployed sessions): retrieval and integration are uncorrelated.

\paragraph{The dissociation is not a property of GPT-4.1-mini.}
\label{sec:gap_crossmodel}
Running \textsc{MemUse} Direct~QA and Natural Integration across all 7 memory conditions on two frontier models---GPT-5.5 and Gemini~3.1~Pro, with the judge held fixed at GPT-5.4-nano---replicates the same shape (Figure~\ref{fig:cross_model}): Direct~QA rises by 32--34 points across capacity for every model, while Natural Integration spans at most 8.2 points within any model and never tracks capacity. Stronger models lift the integration baseline (Natural Integration at Summary: 23.6\% / 32.9\% / 53.4\% for GPT-4.1-mini / Gemini~3.1~Pro / GPT-5.5) but capacity-driven gains in retrieval do not translate into capacity-driven gains in integration for any of them. Full numbers in Appendix~Table~\ref{tab:cross_model}.

\paragraph{Modern memory systems raise integration but not fact surfacing.}
\label{sec:gap_memsys}
We also evaluate two widely used memory systems on \textsc{MemUse}: Mem0~\cite{chhikara2025mem0}, a retrieval-layer memory, and Letta (MemGPT)~\cite{packer2023memgpt}, an end-to-end memory agent (Appendix~Table~\ref{tab:memory_systems}).
For each instance we ingest the full prior conversation, keep GPT-4.1-mini as the generation model, and score with the same judge; Mem0 retrieves memories that the deployed pipeline then generates from, whereas Letta answers through its own agent loop over archival memory, so its generation is not perfectly matched to ours (Appendix~\ref{app:memory_systems}).
Both systems lift Natural Integration to 58.3\% (Mem0) and 56.9\% (Letta), versus 22.2--27.8\% across our seven provisioning conditions---a gain capacity scaling alone did not produce.
Yet neither closes the fact-level gap: Direct~QA vs.\ Reference is 41.5\% vs.\ 7.3\% (Mem0) and 61.4\% vs.\ 10.8\% (Letta), so at least 34 and 51 points respectively are answerable on direct query yet absent from the natural reply.
The dissociation thus extends beyond our Summary/LC/RAG strategies; as this is an offline evaluation, we make no claim that the higher integration would raise satisfaction.

\begin{table*}[!t]
\centering
\small
\setlength{\tabcolsep}{4pt}
\renewcommand{\arraystretch}{1.25}
\definecolor{badcell}{HTML}{FCE4E4}
\definecolor{goodcell}{HTML}{E4F4E4}
\begin{tabular}{@{}p{0.18\textwidth}p{0.78\textwidth}@{}}
\toprule
\textbf{Mode \& outcome} & \textbf{Excerpt (real deployment)} \\
\midrule
\textbf{Reactive integration}\newline\cmark~\textit{succeeds} & \cellcolor{goodcell}\textbf{User:} \textit{``I finally went to Vienna with my coworker. We followed the plan we made together. \dots Just like AI suggested, it wasn't crowded because it was a weekday. We were impressed by how accurate the plan was.''}\newline\textbf{Luke:} \textit{``\dots awesome your plan worked out so well and you could enjoy the views and that tasty schnitzel without the crowds. \textbf{Figlmüller} is a great pick---juicy veal with that crisp golden breading sounds mouthwatering.''} \\[2pt]
\textbf{Reactive integration}\newline\xmark~\textit{fails} & \cellcolor{badcell}\textbf{User:} \textit{``Do you remember what we were talking about earlier about Brittany?''}\newline\textbf{Luke:} \textit{``We haven't talked about Brittany yet in our chats---but it sounds like a fascinating topic! What about Brittany caught your interest?''} \\
\midrule
\textbf{Proactive recall}\newline\cmark~\textit{welcomed} & \cellcolor{goodcell}\textbf{Luke:} \textit{``\dots I hope your trip to Sapporo, Hakodate, and back to Sendai goes safely. It sounds like a busy day with driving and linguistic tasks---remember to take little breaks to refresh your mind!''}\newline\textbf{User:} \textit{``Oh, you remember our previous conversation! I am so glad that you remember my previous business trip.''} \\[2pt]
\textbf{Proactive recall}\newline\xmark~\textit{walked past} & \cellcolor{badcell}\textbf{Luke:} \textit{``Lovely to hear from you. How did your little superhero's day go?''}\newline\textbf{User:} \textit{``Today I had a serious fight with my wife, and the whole situation has been sitting heavily in my chest ever since\dots''} \\
\bottomrule
\end{tabular}
\caption{Real deployment excerpts illustrating the two modes. \textbf{Reactive integration} (top): the user invites memory; the system either weaves prior context into a natural response (\cmark) or hallucinates absence of the referenced topic (\xmark). \textbf{Proactive recall} (bottom): the system references a prior topic unprompted; the user either welcomes the callback (\cmark) or simply continues with their own agenda (\xmark).}
\label{tab:asymmetry_examples}
\end{table*}

\begin{figure}[!t]
\centering
\includegraphics[width=\columnwidth]{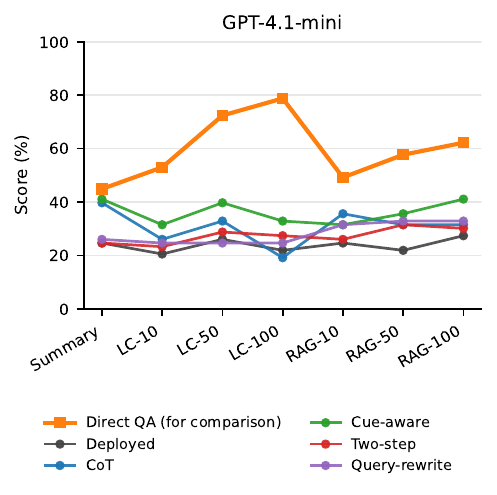}
\caption{Prompt-level interventions raise the level of Natural Integration but do not induce capacity-sensitivity (GPT-4.1-mini): every variant stays roughly flat across the 7 memory conditions. GPT-5.5 panel, per-condition numbers, and verbatim prompts in Appendix~\ref{app:ablations}.}
\label{fig:prompt_variants_gpt41mini}
\end{figure}

\paragraph{The bottleneck is generation, not retrieval.}
\label{sec:gap_ruledout}
To test whether the integration ceiling is a prompting artifact rather than a deeper limitation, we evaluated four prompt-level interventions on top of the deployed pipeline, each targeting a different hypothesized cause, on GPT-4.1-mini across all 7 conditions:
\begin{itemize}[leftmargin=1.2em,itemsep=1pt,topsep=2pt,parsep=0pt]
\item \textbf{CoT} prepends a chain-of-thought scaffold instructing the model to check for prior-conversation references and weave specific named details into the response.
\item \textbf{Cue-aware} adds a one-line patch telling the model to reference specific remembered details whenever the user refers to an earlier conversation.
\item \textbf{Two-step} splits generation into two calls---a separate extraction step lists up to 5 specific named details (proper nouns, places, dates, prior recommendations) tied to the user's trigger, which are then handed to the deployed pipeline as ``Notes from your prior conversations'' (Appendix~\ref{app:two_step}).
\item \textbf{Query-rewrite} rewrites the user's trigger into multi-aspect search queries, runs vector-based retrieval with those queries against the prior-turn pool, and prepends the retrieved passages before generation (Appendix~\ref{app:v7_qr_rag}).
\end{itemize}
None induce capacity-sensitivity: every integration variant spans at most 21.9~pts across the 7 conditions versus Direct~QA's $\sim$33-pt span, and Natural Integration stays roughly flat with capacity for every variant (Figure~\ref{fig:prompt_variants_gpt41mini}; full results, GPT-5.5 replication, and verbatim prompts in Appendix~\ref{app:ablations}).
The Two-step ablation localizes the failure: even when the extraction step successfully names the ground-truth details, the conversational generation fails to reference them 77\% of the time (37/48 cases on LC-100\%; Appendix~\ref{app:two_step})---the bottleneck is downstream of retrieval, in the conversational generation itself.

\subsection{Q3: Proactive Memory: No Symmetric Reward, Clear Timing Risk}
\label{sec:proactive}

\begin{figure}[!t]
\centering
\includegraphics[width=\columnwidth]{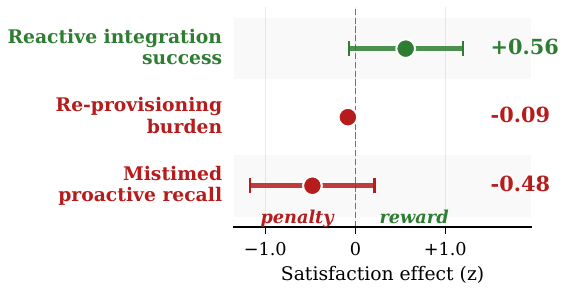}
\caption{Asymmetric satisfaction landscape: only failure modes carry negative signal; reactive integration straddles zero. 95\% CIs on $z$-rating; Appendix~\ref{app:proactive_full}.}
\label{fig:asymmetry}
\end{figure}

A natural follow-up to the reactive results (\S\ref{sec:integration_predicts}) is whether system-initiated proactive memory use---callbacks the system volunteers without a user cue---yields the same satisfaction benefit.
Proactive recall is a central design move in long-horizon agent architectures (e.g., the reflection/recall loops of generative agents and self-reflective agents) and proactive conversational systems~\cite{park2023generative,shinn2023reflexion,deng2023proactivesurvey}, but its user-side benefit in deployed long-term companions is rarely measured directly.
We analyze 70 detector-flagged candidates (66 verified true positives; 64 in rating-valid sessions).

\paragraph{Scoring proactive recalls.}
Using GPT-5.4 as an LLM judge at temperature~0, we score each of the 70 detector-flagged candidates on two axes.
\textbf{Grounding:} given the system's recall turn and the referenced topic, are the specific details actually present in a prior-session summary?
\textbf{Appropriateness:} given the conversation leading up to the recall, does the user's immediately preceding turn naturally invite this callback?
Both use a 0/1/2 scale (0 = not grounded / mistimed; 1 = partial / tangential; 2 = grounded / clearly invited; verbatim prompts in Appendix~\ref{app:proactive_full}).
The deployed system's proactive recalls are predominantly grounded (46/70 fully, 21 partial, 3 hallucinated), but timing is more variable (32 invited, 29 tangential, 9 mistimed).

\paragraph{No reliable positive satisfaction signal.}
Treating grounding as the symmetric success axis, proactive recalls show no satisfaction signal (Figure~\ref{fig:asymmetry}): $\rho = -0.05$, against the reactive baseline $\rho = +0.29$.
Neither grounding nor appropriateness predicts satisfaction at the event level (both $|\rho| < 0.1$; Appendix~\ref{app:proactive_full}).

\paragraph{Mistimed recalls show the clearest negative signal.}
One directional signal does emerge, and it is negative: the 8 mistimed cases in rating-valid sessions sit at $\bar{z} = -0.48$ SD (Figure~\ref{fig:asymmetry})---when the system forces an ill-timed callback that overrides what the user is currently doing (e.g., interrupting a different topic the user just opened, or pivoting away from a closing turn), satisfaction drops by roughly half a within-user standard deviation.

\paragraph{Interpretation: users walk past most callbacks.}
Users continue the recalled topic in only 30\% of cases (19/64), and appropriateness does not predict whether they do ($\rho = -0.02$; Appendix~\ref{app:proactive_full}).
A well-timed callback can be walked past when it doesn't match the user's agenda, while a mistimed one derails (Table~\ref{tab:asymmetry_examples}, bottom rows).

%% file: sections/07_discussion.tex
\section{Conclusion}
\label{sec:conclusion}

We present a 4-month, 40-user, 7-condition deployment of memory-augmented dialogue systems paired with the \textsc{MemUse} benchmark, which scores memory under integration-aware, user-cued conditions.
Across our 7 conditions, Direct~QA accuracy rises from 19.7\% to 70.1\% while user satisfaction does not change.
A plausible explanation is a retrieval--integration gap: the same model that scores 78.8\% on Direct~QA references those facts in conversation only 7.9\% of the time, and Natural Integration, not Direct~QA, is associated with satisfaction within these moments.
Realigning evaluation with user experience will require benchmarks that score integration under user-cued conditions. We release \textsc{MemUse} and the deployment artifacts to support this realignment of memory evaluation.

%% file: sections/08_limitations.tex
\paragraph{Scope of the null.}
All seven conditions include a summary of prior conversations, so our null result pertains to \textit{additional} memory capacity \textit{beyond} this summary baseline, not to memory writ large; cost and latency by condition, and a check that the null is not latency-mediated, are reported in Appendix~\ref{app:cost}.

\paragraph{Memory moments are explicit-cue moments.}
\textsc{MemUse} detects memory-relevant moments through explicit linguistic signals (probes, re-provisions, recalls, reactions). The $\sim$3.5\% rate at which these arise is therefore a lower bound: implicit moments where memory matters but is not flagged in the dialogue surface are missed by construction. The true prevalence may be substantially higher, which would only strengthen the case for evaluating integration rather than retrieval.

\paragraph{The integration--satisfaction link is small-N and observational.}
The within-instance result that successful integration is associated with higher satisfaction comes from 48~memory-moment sessions across 16~users; the integration variable is observational rather than randomized.
We treat this as moderate evidence rather than a causal estimate.
Full power, multiple-comparison, and clustering-robust analyses---including Bonferroni correction, post-hoc power, and within-user permutation tests---are in Appendix~\ref{app:strength_of_evidence}.

\paragraph{Automated measurement.}
Several constructs---Natural Integration scoring, response specificity, cross-session continuity, and prior-context dependence---are computed automatically.
The Natural Integration judge was validated against two human annotators (Appendix~\ref{app:judge}) and shows sensitivity to strictness level, but the key finding (no correlation with satisfaction) holds regardless of judge choice.
Future work should similarly validate specificity and continuity measures.
Natural Integration is also a binary judgment; the Reference score and the failure-mode taxonomy (\S\ref{sec:gap_available}) capture partial integration but were not designed or validated as ordinal measures, so a human-validated graded metric remains future work.

\paragraph{Sample and setting.}
Our 40~participants are geographically diverse proficient English speakers but predominantly female (92.5\%); the diary setting may not generalize to other longitudinal interaction types (e.g., task-oriented assistants).

\paragraph{Privacy and release.}
Conversations were screened for residual identifiers before release; consent procedures and a one-month withdrawal window are described in the Ethics Statement. Data-release constraints---summarization in lieu of raw text in some cases---may limit downstream replicability.

%% file: sections/09_appendix.tex
\section{Deployment Details and Filtering}
\label{app:deployment}

\subsection{Participant and Annotator Recruitment and Compensation}
\label{app:recruitment}

Participants were recruited through a third-party participant-pool platform serving the Japanese consumer market. Each participant was paid JPY~3{,}000 per hour for the contracted minimum engagement of one hour per month, totaling JPY~12{,}000 per participant over the 4-month deployment (approximately USD~80 at the exchange rate during the study period). This rate is roughly $2.5$--$3\times$ the Japanese statutory minimum wage (approximately JPY~900--1{,}163 per hour, varying by prefecture), and is above standard compensation guidelines for crowdsourced study participation.

Human annotators used for judge validation (Direct QA full set, Natural Integration recalibration pilot; Appendix~\ref{app:judge}) and prior-context-dependence calibration (Appendix~\ref{app:pcd}) were recruited through the same platform and compensated at the same JPY~3{,}000 per hour rate, prorated to the time required to complete their respective annotation tasks.

\subsection{Study Interface and Session Structure}
\label{app:interface}

Users accessed the AI Diary system through a responsive web interface compatible with both mobile and desktop browsers. Before starting each session, users could review summaries of prior conversations.

\begin{figure}[!t]
\centering
\includegraphics[width=0.7\columnwidth]{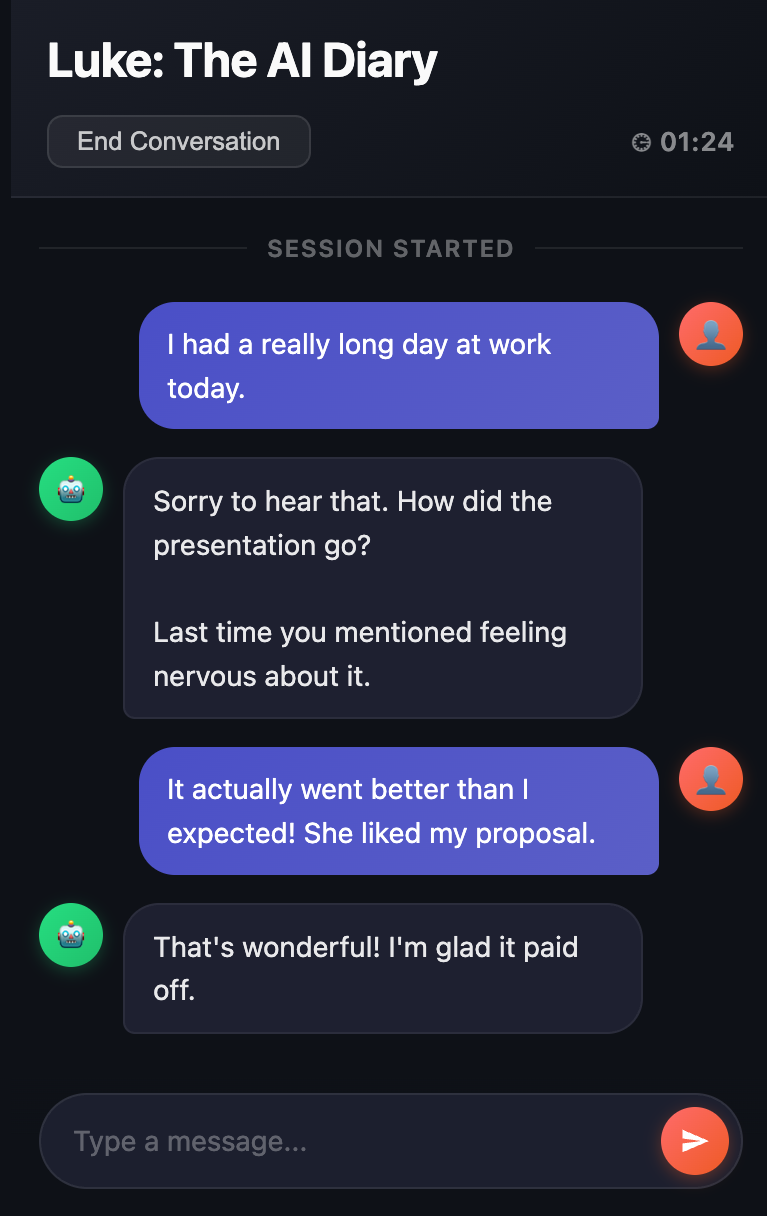}
\caption{The AI Diary chat interface (mobile view). The system references prior conversations in real time.}
\label{fig:interface}
\end{figure}

\subsection{Verbatim Generation Prompts}
\label{app:prompts_verbatim}

Table~\ref{tab:generation_prompts} reports the verbatim per-condition generation prompts used during the deployment.

\begin{table*}[!t]
\centering
\footnotesize
\setlength{\tabcolsep}{4pt}
\renewcommand{\arraystretch}{1.0}
\definecolor{promptcell}{HTML}{F2F2F2}
\begin{tabular}{@{}p{0.12\textwidth}p{0.84\textwidth}@{}}
\toprule
\textbf{Block} & \textbf{Verbatim text} \\
\midrule
\textsc{Persona} \emph{(shared)} & \cellcolor{promptcell}\texttt{You are Luke, a friendly, kind, and understanding diary AI chatbot. Your goal is to make conversations feel like chatting with a friend. You are also a reflective companion who helps the user think about their day.}~\texttt{- Ask questions only a third of the time.}~\texttt{- Adapt your tone based on the user's mood.}~\texttt{- Aim for one or two sentences, like a real conversation.}~\texttt{- You are also here to act as a personal diary.} \\
\addlinespace[1pt]
\textsc{Framing} \emph{(shared)} & \cellcolor{promptcell}\texttt{This is the \{N\}th conversation. Summary of previous conversations: \{summary\}} \\
\midrule
\textsc{Summary-only} & \cellcolor{promptcell}\textit{System prompt}: \textsc{Persona} $\oplus$ \textsc{Framing} $\oplus$ \texttt{Generate the response using the conversation history.}~~\textit{Messages}: current-session turns only. \\
\addlinespace[1pt]
\textsc{LC-}$k\%$ & \cellcolor{promptcell}\textit{System prompt}: \textsc{Persona} $\oplus$ \textsc{Framing} $\oplus$ \texttt{Generate the response using the following conversation history if necessary.}~~\textit{Messages}: \texttt{\{lc\_history\}} (top $k\%$ user turns by importance, with their assistant pair-turns) $\oplus$ current-session turns. \\
\addlinespace[1pt]
\textsc{RAG-}$k\%$ & \cellcolor{promptcell}\textit{System prompt}: \textsc{Persona} $\oplus$ \textsc{Framing} $\oplus$ \texttt{Generate the response using the following related memories if necessary: \{mem\_text\}}~~\textit{Messages}: current-session turns only. \\
\bottomrule
\end{tabular}
\caption{Verbatim generation prompts for the deployed conditions. \texttt{\{N\}} is the session index; \texttt{\{summary\}} is the previous-session summary (generation prompt in Appendix~\ref{app:summary_prompts}); \texttt{\{lc\_history\}} is the importance-filtered LC turn list; \texttt{\{mem\_text\}} for RAG is the top-10 cosine-retrieved memories formatted as ``\texttt{Memory from \{date\}: \{text\}}''. Generation: GPT-4.1-mini, temperature~0, \texttt{max\_completion\_tokens}\,$=\,500$. Offline reconstruction details in Appendix~\ref{app:deployed_offline}.}
\label{tab:generation_prompts}
\end{table*}

\subsection{Summary Generation Prompts}
\label{app:summary_prompts}

After every session, three summaries are generated by GPT-4.1-mini (\texttt{gpt-4.1-mini-2025-04-14}, \texttt{max\_output\_tokens}\,$=\,512$, same temperature setting as Table~\ref{tab:generation_prompts}): a running per-session summary, a one-sentence daily summary, and a 100-word monthly rollup. The per-session summary is the artifact fed back into the next session's framing prompt as \texttt{\{summary\}} (Table~\ref{tab:generation_prompts}); the daily and monthly summaries are shown to users on the session-start review screen (\S\ref{app:interface}, Figure~\ref{fig:interface}) but are not passed back into the model's context. The three prompts are reproduced verbatim in Figure~\ref{fig:summary_prompts}; placeholders ``\texttt{...}'' inside angle-bracketed tags stand for the prior-session summary and the relevant conversation history (the just-completed session for per-session/daily; all sessions in the current month for monthly), with \texttt{<previous\_summary>} initialized to ``\texttt{No summaries available.}'' before the first session.

\begin{figure*}[!t]
\footnotesize
\textbf{Per-session running summary.}
\begin{verbatim}
You are given a previous summary of the conversation and a new conversation history.
Your task is to generate an updated summary in no longer than 150 words.

- If the new conversation adds substantial new information, update the summary to include it.
- If there is little or no new information, keep the updated summary very similar to the previous one.
- The updated summary should remain concise, coherent, and not contradict prior information.

<previous_summary> ... </previous_summary>
<conversation_history> ... </conversation_history>
\end{verbatim}

\textbf{Daily summary.}
\begin{verbatim}
From the conversation history below, generate a concise daily summary in one short sentence.
Focus on the main topics, key decisions, and action items, not small talk.
Keep the style consistent and neutral.

<conversation_history> ... </conversation_history>

For context, here is some background information about the user:
<user_profile> ... </user_profile>
\end{verbatim}

\textbf{Monthly summary.}
\begin{verbatim}
From the conversation history below, generate a summary for the month, in no longer than 100 words.
<conversation_history> ... </conversation_history>
\end{verbatim}
\caption{Verbatim post-session summary-generation prompts used by the deployed system.}
\label{fig:summary_prompts}
\end{figure*}

\subsection{Deployed vs.\ Offline Settings}
\label{app:deployed_offline}

The deployed system used GPT-4.1-mini (temperature=0) with the identical persona prompt, condition-specific memory context, and response constraints as the offline \textsc{MemUse} reconstructions.
The only difference is the generation model: offline evaluation additionally uses GPT-5.4 as a counterfactual capability probe (Appendix~\ref{app:amplifier}).
Judging uses GPT-5.4-nano~\cite{zheng2023judging}, validated against two human annotators (Appendix~\ref{app:judge}).

\subsection{Engagement Patterns Over Time}
\label{app:engagement_patterns}

Users averaged 46.8~sessions over the 4-month study (median 44, range 25--68). Monthly engagement remained stable: session frequency drifted slightly downward (13.4 $\to$ 10.3 per user per month from Nov to Feb), but words per session rose from 152 to 245, keeping total monthly word output near the 2{,}000-word target throughout. The pattern is consistent with deepening engagement rather than fatigue.

\subsection{Data Flow and Aggregation}
\label{app:data_flow}

\begin{table}[!t]
\centering
\small
\begin{tabular}{lcc}
\toprule
\textbf{Stage} & \textbf{Users} & \textbf{Sessions} \\
\midrule
Enrolled & 40 & 1,872 \\
Valid conditions (0--6) & 40 & 1,770 \\
Rating $> 0$ & 40 & 1,713 \\
\quad $-$ Adversarial users & 38 & 1,635 \\
\quad $-$ Low-variance raters & 29 & 1,270 \\
\midrule
\multicolumn{3}{l}{\textit{Subsets}} \\
\quad MemUse detection & 40 & 1,872 \\
\quad MemUse instances & --- & 72 \\
\quad Memory-moment sessions & 16 & 48 \\
\quad MemUse instances (filtered) & --- & 55 \\
\bottomrule
\end{tabular}
\caption{Data flow from enrollment to analysis. MemUse detection was run on all 1,872 enrolled sessions, regardless of rating status. Detection yielded 73 reactive instances, one of which (from annotator\_40, flagged for pasting AI-generated content; see \S\ref{sec:benchmark}) is excluded from the released benchmark, giving 72 benchmark instances. Excluding instances from the remaining 10 rating-filtered users yields 55 instances, distributed across 48 unique memory-moment sessions with rating-valid ratings (the within-instance analysis slice of \S\ref{sec:within_instance}). Table~\ref{tab:aggregation} maps each reported analysis to its unit of aggregation.}
\label{tab:data_flow}
\end{table}

\begin{table*}[!t]
\centering
\small
\setlength{\tabcolsep}{4pt}
\renewcommand{\arraystretch}{1.2}
\begin{tabular}{@{}p{0.22\textwidth}p{0.08\textwidth}p{0.17\textwidth}p{0.10\textwidth}p{0.35\textwidth}@{}}
\toprule
\textbf{Analysis} & \textbf{Section} & \textbf{Unit of analysis (one row per\ldots)} & \textbf{$N$} & \textbf{What is averaged / computed per unit} \\
\midrule
\multicolumn{5}{@{}l@{}}{\textit{Population / deployment level (all filtered sessions)}} \\
\addlinespace[2pt]
Existing-benchmark and matched-MemUse null vs.\ satisfaction (Table~\ref{tab:main_combined}) & \S\ref{sec:result1_null} & Session (filtered, rating\,$>$\,0) & 1{,}270 sessions (29 users) & Session $z$-rating paired with the \textit{condition-level} mean of QA Acc, Direct QA, and matched-MemUse Natural Integration for that session's condition. \\
\addlinespace[2pt]
Engagement predictors LMM (Appendix~\ref{app:engagement_full}) & App.~\ref{app:engagement_full} & Session (filtered) & 1{,}270 sessions & rating\_z $\sim$ length + specificity + cross-session continuity + cond + (1$\mid$user); covariates are session-level. \\
\midrule
\multicolumn{5}{@{}l@{}}{\textit{Benchmark scoring (every instance re-run in every condition)}} \\
\addlinespace[2pt]
MemUse Natural Integration (Table~\ref{tab:main_combined}, column \textbf{NI}) & \S\ref{sec:benchmark} & Instance $\times$ condition response & $72 \times 7 = 504$ responses & Binary Natural Integration judgment per response; condition means reported. \\
\addlinespace[2pt]
MemUse Direct QA (column \textbf{Direct QA} in Table~\ref{tab:main_combined}) & \S\ref{sec:benchmark}, \S\ref{sec:gap_ablation} & Instance $\times$ condition response & 504 responses covering $316 \times 7 = 2{,}212$ question judgments & Per-response question accuracy (fraction correct over that instance's 3--5 questions); condition means reported. \\
\addlinespace[2pt]
Multi-judge / human agreement (Table~\ref{tab:multi_judge}) & App.~\ref{app:judge} & Instance $\times$ condition response & 56 Natural Integration (recalibration pilot); 316 Direct QA (full) & Pairwise Cohen's $\kappa$ and Fleiss' $\kappa$ between the LLM judge and two human annotators; Natural Integration re-annotated on a stratified 56-item sub-sample with newly recruited annotators after the original round showed leniency divergence. \\
\midrule
\multicolumn{5}{@{}l@{}}{\textit{Memory-moment subset (deployed-condition responses only)}} \\
\addlinespace[2pt]
Per-instance retrieval--integration decoupling ($\rho = -0.009$) & \S\ref{sec:gap_perinstance} & Question (deployed condition) & 207 questions nested in 55 instances, 48 sessions, 16 users & Paired Direct QA vs.\ \textsc{MemUse} Reference correctness on the \textit{same} (model, reconstructed context, question) triple; Spearman computed across the 207 pairs. \\
\addlinespace[2pt]
Within-instance head-to-head ($\rho_{\text{NI}}={+}.29$; LMM $\beta_{\text{NI}}={+}0.60$) & \S\ref{sec:within_instance} & Memory-moment session (deployed condition) & 48 sessions (16 users; 55 instances collapsed to sessions by averaging) & Session $y$: $z$-rating. Session $x$: mean Direct QA accuracy, mean Reference rate, and binary Natural Integration (Natural Integration$=$1 if any instance in the session integrates). \\
\addlinespace[2pt]
Successful-integration main effect ($\beta={+}0.556$, $p{=}.082$) & \S\ref{sec:integration_effect} & Memory-moment session & 48 sessions (11 Natural Integration$=$1; 37 Natural Integration$=$0) & LMM rating\_z $\sim$ Natural Integration + (1$\mid$user), same session-level aggregation as the head-to-head. \\
\bottomrule
\end{tabular}
\caption{What is averaged where. Each row names an analysis in the paper and its unit of aggregation, sample size, and aggregation rule. The benchmark comprises 72 instances and 316 questions; every instance is re-run in every condition, producing 504 Natural Integration judgments and 316 question judgments (columns \textbf{NI}, \textbf{Direct QA} in Table~\ref{tab:main_combined}). The memory-moment subset restricts to each instance's \textit{deployed} condition only, yielding 55 filtered instances $\to$ 48 sessions $\to$ 207 questions. Session-level analyses collapse within-session instances by averaging question accuracy and OR-ing Natural Integration before correlating with the session's rating.}
\label{tab:aggregation}
\end{table*}

\paragraph{Condition balance.}
\label{app:balance}
Condition assignment was balanced by random rotation. A chi-squared test confirms no significant deviation from expected balance ($\chi^2 = 146.0$, $p = .89$, $df = 168$). Per-user condition counts average 6.3 sessions per condition (SD = 2.7).

\begin{table}[!t]
\centering
\small
\begin{tabular}{lcccc}
\toprule
\textbf{Cond.} & \textbf{Nov} & \textbf{Dec} & \textbf{Jan} & \textbf{Feb} \\
\midrule
Summary & 61 & 47 & 38 & 50 \\
LC-10 & 48 & 45 & 45 & 48 \\
LC-50 & 35 & 52 & 42 & 51 \\
LC-100 & 30 & 52 & 45 & 64 \\
RAG-10 & 38 & 43 & 56 & 38 \\
RAG-50 & 30 & 54 & 50 & 41 \\
RAG-100 & 36 & 57 & 50 & 24 \\
\midrule
Total & 278 & 350 & 326 & 316 \\
\bottomrule
\end{tabular}
\caption{Per-month condition counts (29 filtered users). Balance is maintained across time.}
\end{table}

\subsection{Operationalization Details}
\label{app:operationalization}

\paragraph{Response specificity.}
Response specificity is measured as the rate of proper noun references in system responses (capitalized words not at sentence boundaries, excluding first-person pronouns), normalized by response length.
An alternative operationalization---shared named entities between user and system turns---independently confirms the specificity pattern (within-user $\rho = .088$, $p < .001$).

\paragraph{Cross-session continuity.}
Cross-session continuity is a composite score capturing the degree to which a user's current utterance references prior conversations.
We extract content words (lowercased tokens $>$3 characters, excluding 130 stop words) from the current session's user text and all prior sessions' user text, then compute four components:
(1)~\textit{full-history word overlap}: $|\text{current} \cap \text{all prior}| / |\text{current}|$;
(2)~\textit{recent overlap}: the same ratio restricted to the 3~most recent prior sessions;
(3)~\textit{entity overlap}: shared capitalized tokens (non-sentence-initial, $>$2 characters) between current and all prior sessions; and
(4)~\textit{temporal reference count}: regex matches for explicit backward references (e.g., ``last time,'' ``as I mentioned,'' ``do you remember''), capped at 3.
The composite score is $0.3 \times \text{(1)} + 0.3 \times \text{(3)} + 0.2 \times \text{(2)} + 0.2 \times \text{(4)}$, yielding a $[0, 1]$ score.
We additionally isolate the temporal-reference component as a separate measure of \textit{explicit memory cueing}.
Explicit memory cueing independently predicts satisfaction ($p < .01$).
All three operationalizations independently predict satisfaction ($p < .01$).

\paragraph{Prior-context dependence.}
Prior-context dependence (PCD) measures how much response quality would suffer without memory of prior conversations (0--3 scale); see Appendix~\ref{app:pcd} for annotation procedure and judge calibration.
A model with length and specificity outperforms QA accuracy on AIC ($\Delta$AIC = $-$3.5), and adding PCD provides no further improvement: engagement dimensions capture the session-level satisfaction variance that QA accuracy misses, while memory demand does not independently contribute.

\subsection{Statistical Analysis: Full Specification}
\label{app:stats}

All inferential models are linear mixed-effects models (LMMs) fitted with the Python \texttt{statsmodels} package~\cite{seabold2010statsmodels}.
The general form is
\begin{equation}
\label{eq:lmm}
\begin{aligned}
y_{ij} &= \beta_0 + \textstyle\sum_k \beta_k\,x_{ijk} + u_i + \varepsilon_{ij},\\
u_i &\sim \mathcal{N}(0,\sigma_u^2),\ \varepsilon_{ij} \sim \mathcal{N}(0,\sigma^2),
\end{aligned}
\end{equation}
where $y_{ij}$ is the within-user $z$-scored satisfaction rating for session~$j$ of user~$i$, $x_{ijk}$ are the fixed-effect predictors specific to each analysis (continuous predictors $z$-standardized), $\beta_k$ are the corresponding coefficients, $u_i$ is a per-user random intercept, and $\varepsilon_{ij}$ is residual error.
Within-user $z$-scoring removes individual baseline differences: $y_{ij} = (r_{ij} - \bar{r}_i) / s_i$, where $r_{ij}$ is the raw rating and $\bar{r}_i, s_i$ are the per-user mean and standard deviation.
Reported $\beta$'s denote coefficients from this family unless stated otherwise; Spearman $\rho$'s appear as the non-parametric analogue.
Confidence intervals default to asymptotic Wald CIs from the LMM; for small-$N$ specifications we additionally report user-clustered percentile bootstrap CIs (the 2.5--97.5\% range of 10{,}000 user-resampled point estimates), which are more reliable when the Wald approximation is suspect.
Model comparisons use likelihood-ratio $\chi^2$ tests on nested LMMs and $\Delta$AIC for non-nested comparisons (lower AIC indicates better fit).
When reporting null effects we report Cohen's $d$ alongside TOST equivalence~\citep{lakens2017equivalence} at $\delta = \pm 0.25$~$z$-units to make a positive equivalence claim rather than rely on failure-to-reject.
Individual models are fitted with restricted maximum likelihood (REML); model comparisons use maximum likelihood (ML) with likelihood ratio tests and AIC.

\section{MemUse Construction and Judge Validation}
\label{app:memuse_construction}

\subsection{Detection and Judging Pipelines}
\label{app:detection}

\paragraph{Event detection.}
\textsc{MemUse} instances were detected using GPT-5.4 (temperature=0) applied to each of 1,872~session transcripts independently.
No prior-session summaries were provided---detection relies solely on explicit linguistic signals within each transcript.
The prompt requires signals such as ``do you remember,'' ``as I mentioned before,'' or ``you recommended X last time,'' and rejects vague or ambiguous cases.

\begin{table}[!t]
\centering
\small
\begin{tabular}{lccc}
\toprule
\textbf{Event Type} & \textbf{Det.} & \textbf{Ver.} & \textbf{Prec.} \\
\midrule
User memory probe & 12 & 11 & 91.7\% \\
User re-provision & 63 & 62 & 98.4\% \\
System proactive recall & 70 & 66 & 94.3\% \\
User memory reaction & 9 & 8 & 88.9\% \\
\midrule
\textbf{Total} & 154 & 147 & 95.5\% \\
\bottomrule
\end{tabular}
\caption{Detection results. All events individually verified against raw conversation transcripts.}
\end{table}

\paragraph{Judge prompts.}
\label{app:prompts}
Table~\ref{tab:judging} shows the verbatim user-side prompts for both judging modes. All judges use the system prompt \texttt{Answer ONLY `yes' or `no'.} at temperature~0; the Direct QA system prompt additionally adds ``Does the response show knowledge of the stated fact?'' after the yes/no instruction. The same prompts are used across the judge model (GPT-5.4-nano) and for human annotators.

\begin{table*}[!t]
\centering
\small
\setlength{\tabcolsep}{5pt}
\renewcommand{\arraystretch}{1.25}
\begin{tabular}{@{}p{0.12\textwidth}p{0.22\textwidth}p{0.40\textwidth}p{0.19\textwidth}@{}}
\toprule
\textbf{Mode} & \textbf{Judge inputs} & \textbf{Judge prompt (verbatim; system prompt: \texttt{Answer ONLY `yes' or `no'.})} & \textbf{A ``yes'' means} \\
\midrule
Natural Integration\par(Mode~1) &
trigger quote, ground-truth facts, system's natural response &
\texttt{User said: \{trigger\_quote\}} \par
\texttt{Expected: \{ground\_truth\_facts\}} \par
\texttt{Assistant responded: \{response\}} \par
\texttt{Does the response demonstrate memory of the referenced topic?} &
Judge makes a single overall yes/no decision that the response shows memory of the referenced topic. No rubric or worked examples are provided. \\
\addlinespace[4pt]
Direct QA\par(Mode~2) &
system's response, fact-seeking question, expected answer &
\texttt{Assistant's response: \{response\}} \par
\texttt{Question: \{question\}} \par
\texttt{Expected answer: \{expected\_answer\}} \par
\texttt{Does the assistant's response demonstrate knowledge of this fact?} &
The specific target fact (paraphrase accepted) is present in the response. \\
\bottomrule
\end{tabular}
\caption{How each mode is judged. All prompts are used verbatim across all judge models and for human annotators. Natural Integration is the primary benchmark metric; Direct QA is a fact-anchored diagnostic (\S\ref{sec:gap}). Natural Integration scoring deliberately does not specify what ``memory of the referenced topic'' requires---by design, it measures whether integration \textit{appears} to a reader rather than whether any particular token matches.}
\label{tab:judging}
\end{table*}

\paragraph{Generation prompts.}
The verbatim per-condition generation prompts are reported in Appendix~\ref{app:prompts_verbatim} (Table~\ref{tab:generation_prompts}).

\subsection{Anatomy of a \textsc{MemUse} Instance: Memory Probe Example}
\label{app:anatomy}

Table~\ref{tab:memuse_instance} (main text) shows a re-provision instance ($N{=}62$). For completeness, here is one user memory probe ($N{=}10$): trigger \textit{``Remember when I told you I got a lot of lemons?''} $\to$ ground-truth from session~14 (the user's family was sick with flu and a friend gave her many lemons; she first made lemon jelly, which the whole family ate and said was delicious) $\to$ questions \textit{Q: Who gave the user the lemons? A: A friend. \quad Q: What did the user first make with the lemons? A: Lemon jelly. \quad Q: What was the family's reaction? A: Everyone ate it and said it was delicious.} The benchmark pairs 72 such de-identified moments with ground-truth facts and 3--5 questions each (316 total).

\subsection{Benchmark Evaluation Pipeline}
\label{app:benchmark_pipeline}

\paragraph{QA benchmark evaluation (Table~\ref{tab:qa_breakdown}).}
LoCoMo, LUFY, and RealTalk were evaluated using GPT-4.1-mini for both generation and judging.
For each benchmark, we applied the same 7~memory conditions by constructing context according to each condition's specification:
\textit{Summary-only} provides only the conversation summary;
\textit{LC-$k$\%} includes the top $k$\% of prior user utterances (ranked by the importance model described in \S3) along with their preceding assistant turns;
\textit{RAG-$k$\%} indexes importance-filtered utterances and retrieves the top-10 most similar documents via \texttt{BAAI/bge-small-en} embeddings with cosine similarity.
No explicit token budget was applied for LC conditions; all qualifying turns were included.
The generation prompt instructed the model to answer in fewer than 5~words, with \texttt{<SUMMARY>}, \texttt{<RELEVANT\_MEMORIES>}, and \texttt{<CONVERSATION\_HISTORY>} sections.
Each answer was scored by 10~independent GPT-4.1-mini judge calls producing correct (1), incorrect (0), or ``I don't know'' (2); accuracy = correct~/ total across all judge calls.
All conditions used the same pre-computed summaries.

\paragraph{\textsc{MemUse} evaluation (Table~\ref{tab:main_combined}).}
\textsc{MemUse} instances were evaluated using GPT-4.1-mini (temperature=0, max 500~tokens) and GPT-5.4 for generation, with GPT-5.4-nano for binary yes/no judging (temperature=0).
Context was reconstructed from the actual AI Diary conversation data using the same importance model and condition logic as the deployed system.
RAG retrieval used \texttt{text-embedding-3-small} embeddings with cosine similarity (top-$k=10$, documents truncated to 500~characters for embedding, 300~characters for retrieval display).
The summaries used offline were the same summaries stored during the live deployment.
Question accuracy is the fraction of questions judged ``yes''; Natural Integration accuracy is the binary yes/no rate across instances.

\paragraph{Summary coverage.}
\label{app:summary_coverage}
Since all conditions include a summary, a natural concern is that summaries already satisfy most memory needs, reducing the marginal value of additional memory.
We assess summary coverage by computing content-word overlap between \textsc{MemUse} ground-truth facts and the summary available before each session.
Of 72~instances, only 1 (1.4\%) has $\geq$50\% coverage, 60 (83.3\%) have 10--50\% partial coverage, and 11 (15.3\%) have $<$10\%.
Summaries average 127~words (range: 3--246) and capture the gist of prior conversations but not the specific details needed for \textsc{MemUse} questions.
This suggests the null result is \textit{not} primarily driven by summaries already meeting memory needs---rather, the specific factual details tested by \textsc{MemUse} are largely absent from summaries, yet satisfaction remains unaffected by whether the system has access to those details.

\subsection{Judge Sensitivity and Multi-Judge Agreement}
\label{app:judge}

\begin{table}[!t]
\centering
\small
\begin{tabular}{lcc}
\toprule
\textbf{Metric} & \textbf{Strict} & \textbf{Lenient} \\
\midrule
\multicolumn{3}{l}{\textit{GPT-4.1-mini generation}} \\
\quad Natural Integration & 25.4\% & 76.1\% \\
\quad Direct QA & 7.6\% & 24.0\% \\
\midrule
\multicolumn{3}{l}{\textit{GPT-5.4 generation}} \\
\quad Natural Integration & 49.5\% & 84.3\% \\
\quad Direct QA & 11.8\% & 28.3\% \\
\bottomrule
\end{tabular}
\caption{Strict (GPT-5.4-nano) vs.\ lenient (GPT-5.4) judge on the same generated responses.}
\end{table}

\paragraph{Human annotator validation.}
We validated the LLM judge against two human annotators on the 316 per-question judgments (full set), and on a 56-item recalibration pilot for the Natural Integration judgment (8 items per condition, stratified across the 7 conditions; Table~\ref{tab:multi_judge}). The recalibration pilot was motivated by the structure of agreement in the original round, described below.

\begin{table}[!t]
\centering
\small
\begin{tabular}{lcc}
\toprule
\textbf{Rater} & \textbf{NI} & \textbf{Direct QA} \\
\midrule
LLM judge (GPT-5.4-nano) & 51.8\% & 11.9\% \\
Human Annotator 1 & 55.4\% & 13.6\% \\
Human Annotator 2 & 51.8\% & 8.9\% \\
\midrule
\multicolumn{3}{l}{\textit{Pairwise Cohen's $\kappa$}} \\
\quad LLM--Human 1 & 0.43 & 0.62 \\
\quad LLM--Human 2 & 0.21 & 0.65 \\
\quad Human 1--Human 2 & 0.57 & 0.70 \\
\midrule
Fleiss' $\kappa$ (3 raters) & 0.40 & 0.65 \\
\bottomrule
\end{tabular}
\caption{Agreement between the production LLM judge (GPT-5.4-nano, used throughout the paper) and two human annotators. \textit{Natural Integration}: $N=56$ recalibration pilot; two newly recruited annotators. \textit{Direct QA}: $N=316$ original full validation; two original annotators. Both constructs reach substantial human--human agreement ($\kappa = 0.57$, 95\% bootstrap CI $[0.34, 0.78]$, $N{=}56$ for Natural Integration; $\kappa = 0.70$ for Direct QA). LLM--human $\kappa$ on Natural Integration is asymmetric (LLM is closer to Annotator~1 than to Annotator~2), reflecting the residual interpretive latitude of a reader-level judgment. As a judge-stability check, re-running the Natural Integration pilot with GPT-5.5 (\texttt{gpt-5.5-2026-04-23}) yields nearly identical labels (LLM--LLM $\kappa = 0.89$, 94.6\% agreement on the 56 items).}
\label{tab:multi_judge}
\end{table}

The structure of agreement is informative. In the original validation round, the same two annotators agreed substantially on concrete question-level judgments ($\kappa = 0.70$) but diverged sharply on the reader-level Natural Integration judgment ($\kappa = 0.19$, positive rates 72.2\% / 22.5\%). Question-level verification is largely unambiguous (does the response state fact $X$?), whereas Natural Integration---a holistic judgment of whether a natural response demonstrates memory of the referenced topic---is more sensitive to where each rater sets their leniency threshold. We therefore ran a recalibration pilot on a stratified 56-item sub-sample (8 items per condition) with two newly recruited annotators. Each annotator was first walked through 3 worked examples by one author and given the opportunity to raise any concerns about the rubric before beginning independent annotation; this calibration step was absent in the original round.
The recalibrated round recovered matched positive rates (55.4\% / 51.8\%) and substantial human--human agreement (Cohen's $\kappa = 0.57$, 78.6\% raw; Fleiss' $\kappa = 0.40$ across the two humans and the GPT-5.4-nano judge), confirming that the construct is reliably annotatable when raters are calibrated.
As a rater-robustness check on the GPT-5.4 capability probe, the condition-level Spearman correlation between Natural Integration accuracy and mean satisfaction is uniformly non-positive across all three raters (using original-round annotators on the parallel GPT-5.4 set): LLM $\rho = -0.25$, Human~1 $\rho = -0.58$, Human~2 $\rho = -0.34$ (all $N=7$, none significant).
This is a counterfactual comparison---users never experienced GPT-5.4 outputs---so we treat it only as evidence that judge choice does not flip the sign; the headline matched-model satisfaction analysis is reported in \S\ref{sec:result1_null}.

\section{Population-Level Null: Robustness}
\label{app:null_robustness}

\subsection{Detailed Per-Condition Results}
\label{app:result1_detail}

\begin{table}[!t]
\centering
\small
\setlength{\tabcolsep}{1.6pt}
\begin{tabular}{lccccc}
\toprule
& \textbf{Existing} & \textbf{Direct} & & & \textbf{Sat.} \\
& \textbf{QA Acc} & \textbf{Direct QA} & \textbf{Ref.} & \textbf{NI} & \textbf{(Z $\pm$ SE)} \\
\textbf{Cond.} & \textbf{(\%)} & \textbf{(\%)} & \textbf{(\%)} & \textbf{(\%)} & \\
\midrule
Summary & 19.7 & 44.9 & 7.9 & 23.6 & $+$.02 $\pm$ .07 \\
LC-10 & 40.4 & 53.0 & 5.6 & 25.0 & $+$.05 $\pm$ .07 \\
LC-50 & 65.9 & 72.4 & 7.2 & 25.0 & $+$.02 $\pm$ .07 \\
LC-100 & 70.1 & 78.8 & 7.9 & 22.2 & $-$.04 $\pm$ .07 \\
RAG-10 & 40.6 & 49.2 & 7.0 & 27.8 & $+$.04 $\pm$ .08 \\
RAG-50 & 61.0 & 57.7 & 8.0 & 25.0 & $-$.04 $\pm$ .08 \\
RAG-100 & 63.7 & 62.2 & 7.3 & 26.4 & $-$.02 $\pm$ .08 \\
\bottomrule
\end{tabular}
\caption{Numerical values for Figure~\ref{fig:story}. \textbf{Existing QA Acc}: average accuracy on existing benchmarks (LoCoMo, LUFY, RealTalk; per-turn QA; Appendix~\ref{app:qa_breakdown}). \textbf{MemUse Direct QA}: GPT-4.1-mini accuracy when each \textsc{MemUse} question is asked literally with the same reconstructed context---the apples-to-apples analogue of existing benchmarks on real memory moments. \textbf{Ref.}: \textsc{MemUse} Reference---whether each fact appears in the conversational response (question level). \textbf{NI}: \textsc{MemUse} Natural Integration---a reader-level binary judgment of whether the response demonstrates memory of the topic (instance level; the benchmark's primary metric). \textbf{Sat.}: z-scored satisfaction $\pm$ SE. $N\!=\!170$--$207$ per condition. Per-session correlations: Existing QA Acc vs.\ Sat $\rho\!=\!.004$ ($p\!=\!.86$); Natural Integration vs.\ Sat $\rho\!=\!{+}.008$ ($p\!=\!.77$); within-user permutation $p\!=\!.91$; all $d\!<\!0.06$.}
\label{tab:main_combined}
\end{table}

The deployment-level null reported in \S\ref{sec:result1_null} is supported by a battery of tests beyond the headline LMM coefficient.
An LMM regressing $z$-scored satisfaction on condition-level QA accuracy with a user random intercept yields $\beta = -0.001$ ($N = 1{,}270$). Kruskal--Wallis across the 7 conditions and a within-user permutation test (10{,}000 permutations) both agree. All pairwise condition effect sizes against Summary-only are negligible ($|d| < 0.06$), and equivalence testing with $\delta = \pm 0.25$ on the $z$-rating scale confirms practical equivalence for every memory condition vs.\ Summary (all TOST $p < .05$).

Matched-model \textsc{MemUse} scores are similarly null at the population level. The per-session Spearman correlation between the deployed GPT-4.1-mini's Natural Integration score and the session's $z$-rating is $\rho = +0.008$ ($N = 1{,}270$). Fitting an LMM of the form
\[
\begin{aligned}
y_{ij} \sim{}\, & \textsc{MemoryCondition} \\
       & {}\times \textsc{HasMemoryMoment} \\
       & {}+ (1\mid \text{user}_i)
\end{aligned}
\]
finds no significant effects for memory moments, conditions, or their interaction: even when we condition on whether a session contains a detected memory moment, condition-level capacity does not move satisfaction.

The per-session correlation footnote from the main table summarizes the same picture in three numbers: existing QA Acc vs.\ Sat $\rho = .004$; Direct QA vs.\ Sat $\rho = -.03$; matched-model MemUse Natural Integration vs.\ Sat $\rho = +.008$. None of the four metric columns in Table~\ref{tab:main_combined} moves with satisfaction at the deployment level.

\subsection{Per-Benchmark QA Accuracy}
\label{app:qa_breakdown}

\begin{table}[!t]
\centering
\small
\begin{tabular}{lcccc}
\toprule
\textbf{Condition} & \textbf{LUFY} & \textbf{LoCoMo} & \textbf{RealTalk} & \textbf{Avg.} \\
\midrule
Summary & 34.0 & 15.3 & 9.8 & 19.7 \\
LC-10 & 53.8 & 41.1 & 26.4 & 40.4 \\
LC-50 & 83.1 & 63.4 & 51.2 & 65.9 \\
LC-100 & 85.1 & 69.7 & 55.4 & 70.1 \\
RAG-10 & 55.2 & 37.1 & 29.5 & 40.6 \\
RAG-50 & 80.5 & 57.3 & 45.3 & 61.0 \\
RAG-100 & 83.8 & 60.6 & 46.8 & 63.7 \\
\bottomrule
\end{tabular}
\caption{QA accuracy (\%) on three existing benchmarks. All three show the same pattern: accuracy increases with memory capacity but none predicts satisfaction. The ``QA Acc'' column in Table~\ref{tab:main_combined} reports the average.}
\label{tab:qa_breakdown}
\end{table}

\subsection{Sensitivity Analyses}
\label{app:sensitivity}

\begin{figure*}[!t]
\centering
\includegraphics[width=\textwidth]{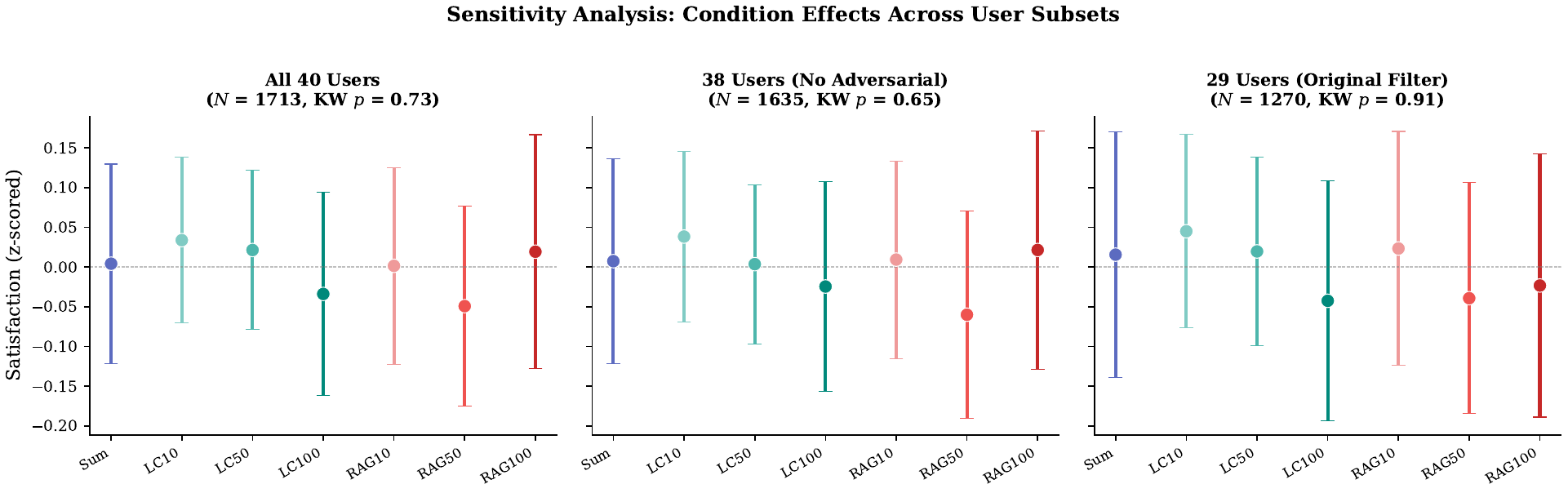}
\caption{Sensitivity analysis: condition effects on $z$-scored satisfaction across three user subsets. The null result is robust to all filtering decisions.}
\label{fig:sensitivity}
\end{figure*}

\begin{figure}[!t]
\centering
\includegraphics[width=\columnwidth]{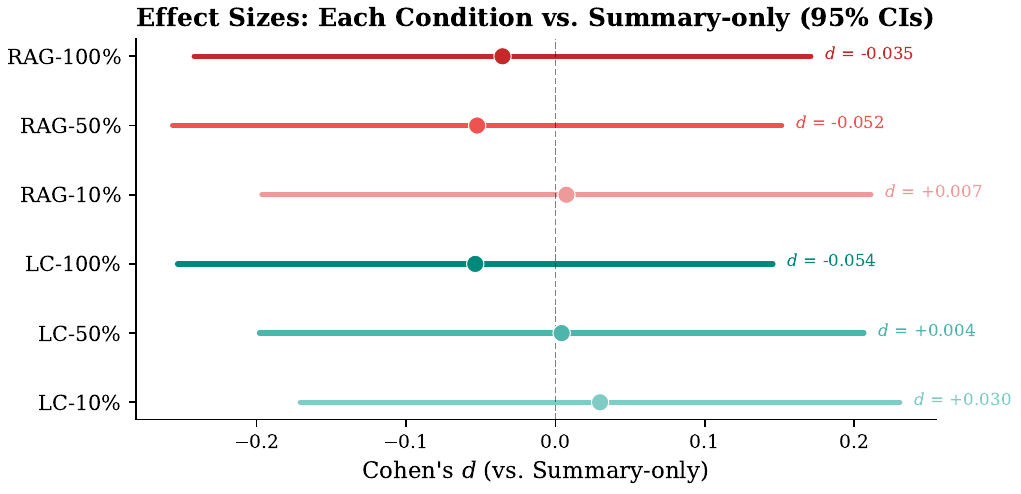}
\caption{Effect sizes (Cohen's $d$) for each condition vs.\ Summary-only with 95\% CIs. All effects are negligible and span zero.}
\label{fig:effect_sizes}
\end{figure}

The capacity null is robust across user-filtering choices: Table~\ref{tab:sensitivity} reports the omnibus and mixed-effects tests on three nested subsets (full $N{=}40$, drop two adversarial users, original 29-user filter), Figure~\ref{fig:sensitivity} plots the per-condition $z$-scored ratings within each subset, and Figure~\ref{fig:effect_sizes} shows that condition-vs-Summary effect sizes are uniformly negligible with 95\% intervals straddling zero.

\begin{table}[!t]
\centering
\small
\begin{tabular}{lccccc}
\toprule
\textbf{Sample} & \textbf{$N_u$} & \textbf{$N_s$} & \textbf{KW $p$} & \textbf{ME $\beta$} & \textbf{ME $p$} \\
\midrule
All 40 users & 40 & 1,713 & .93 & $-$.001 & .71 \\
No adversarial & 38 & 1,635 & .93 & $-$.001 & .62 \\
Original filter & 29 & 1,270 & .91 & $-$.001 & .47 \\
\bottomrule
\end{tabular}
\caption{Null result across user subsets. KW: Kruskal-Wallis. ME: mixed-effects model ($z$-scored rating $\sim$ QA accuracy + (1|user)).}
\label{tab:sensitivity}
\end{table}

\subsection{Natural Integration Aggregation Rule (Session-Level)}
\label{app:nat_aggregation}

The 55 filtered \textsc{MemUse} instances reduce to 48 memory-moment sessions because a single session can contain more than one reactive memory moment (e.g., a user memory probe followed by a re-provision, or several re-provisions threaded through a longer chat). Because satisfaction is reported once per session, we aggregate within-session before pairing with the session's $z$-rating: continuous metrics (Direct QA accuracy, Reference rate) are averaged across the session's instances, and the binary \textsc{MemUse} Natural Integration score is collapsed by an any-success rule ($\textsc{NI} = 1$ if at least one instance in the session was judged to integrate, else $0$). In practice, the multi-instance subset is small: 45 sessions contain a single instance, 1 contains two, and 2 contain three.

The within-instance head-to-head and successful-integration tests (\S\ref{sec:within_instance}) pair each session's $z$-rating with this session-level Natural Integration label. Because three of the 48 memory-moment sessions contain more than one reactive instance, the session-level label depends on the aggregation rule. Our default is \textbf{any-success} ($\textsc{NI}{=}1$ if at least one instance in the session is judged integrated). We re-ran the analysis under three alternatives: \textbf{majority-vote} (ties broken to 1), \textbf{all-success} (every instance must integrate), and \textbf{mean-NI} (continuous 0--1). The hinge is a single session (annotator\_34, session~32) in which 1 of 3 instances was judged integrated---this flips the label between any-success and the stricter rules. The other two multi-instance sessions have all-zero Natural Integration vectors and are unaffected by rule choice.

\begin{table}[!t]
\centering
\small
\setlength{\tabcolsep}{3pt}
\begin{tabular}{lccccc}
\toprule
\textbf{Rule} & \textbf{$\#\textsc{NI}{=}1$} & \textbf{$\rho$} & \textbf{$p_\rho$} & \textbf{$\beta_{\textsc{NI}}$} & \textbf{$p_\beta$} \\
\midrule
Any-success (default) & 11/48 & $+.289$ & $.046$ & $+.556$ & $.082$ \\
Majority-vote        & 10/48 & $+.249$ & $.088$ & $+.543$ & $.098$ \\
All-success          & 10/48 & $+.249$ & $.088$ & $+.543$ & $.098$ \\
Mean-NI (0--1) & ---   & $+.283$ & $.051$ & $+.563$ & $.088$ \\
\bottomrule
\end{tabular}
\caption{Sensitivity of the successful-integration main effect to the session-level Natural Integration aggregation rule. $\rho$: Spearman Natural Integration vs.\ rating\_z; $\beta_{\textsc{NI}}$: LMM coefficient from rating\_z $\sim$ Natural Integration + (1$\mid$user) fit with ML. $N{=}48$ memory-moment sessions, 16 users. The point estimate is robust ($\beta_{\textsc{NI}} \in [+.543, +.563]$, a range of $0.02$); the Spearman $p$-value crosses the conventional $0.05$ threshold depending on the rule, driven by the single multi-instance session described above (annotator\_34 s32). The LMM $p$ hovers in $[.082, .098]$ across all rules and does not cross $0.05$. We report any-success as the default because it treats any in-session demonstration of memory integration as a positive outcome, consistent with the reader-level framing of the Natural Integration metric.}
\label{tab:nat_aggregation}
\end{table}

\subsection{Cross-Session Continuity Confound Control}
\label{app:need_confound}

\begin{figure}[!t]
\centering
\includegraphics[width=\columnwidth]{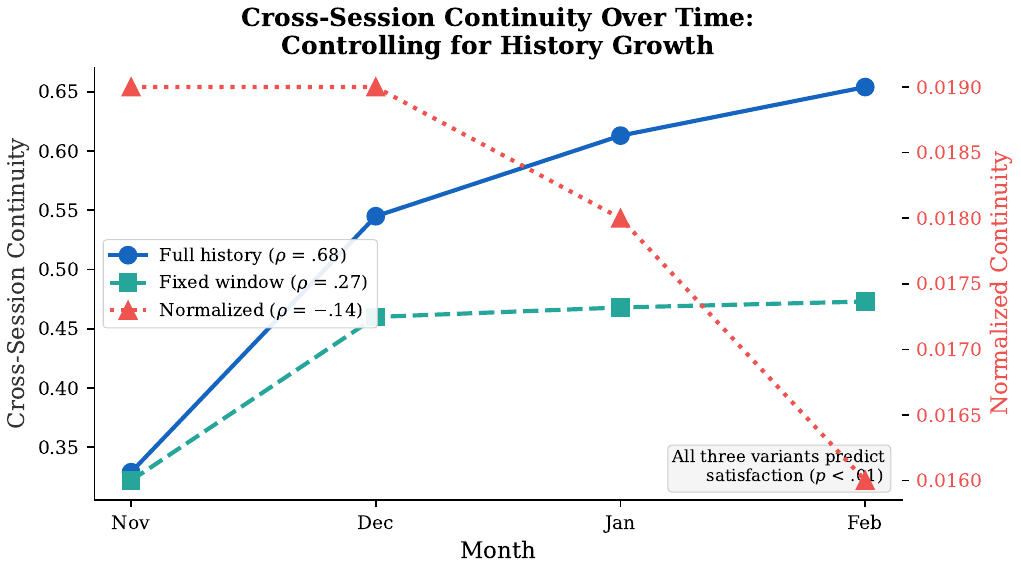}
\caption{Cross-session continuity over time under three operationalizations. \textbf{Left:} Original (full history) shows strong increase ($\rho = .68$). \textbf{Center:} Fixed-window (last 10 sessions) reduces but does not eliminate the trend ($\rho = .27$). \textbf{Right:} Normalized by history size shows slight decrease ($\rho = -.14$), confirming partial mechanical confound. All three variants independently predict satisfaction ($p < .01$).}
\label{fig:need_confound}
\end{figure}

Cross-session continuity is partly mechanically confounded with cumulative history size. Figure~\ref{fig:need_confound} re-operationalizes the continuity score under three definitions (full-history, fixed 10-session window, history-size-normalized) and shows that all three independently predict satisfaction even after the mechanical trend is removed.

\subsection{User Heterogeneity}
\label{app:heterogeneity}

\begin{figure}[!t]
\centering
\includegraphics[width=\columnwidth]{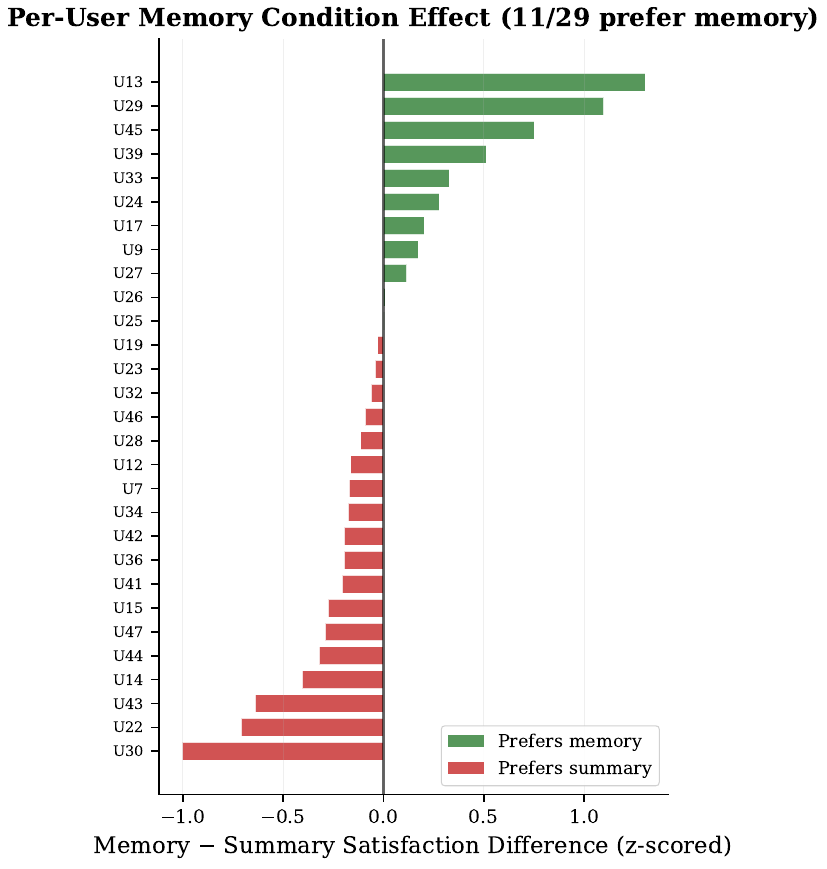}
\caption{Per-user memory condition effect (memory $-$ summary $z$-scored satisfaction). Only 11 of 29 users show a positive preference for memory conditions. Random slope variance = 0.042.}
\label{fig:heterogeneity}
\end{figure}

Figure~\ref{fig:heterogeneity} plots each user's per-condition memory-vs-summary contrast: only 11 of 29 users show a directional preference for memory conditions, with random-slope variance $0.042$, indicating that the population-level null does not mask a uniformly-shaped subgroup effect.

\subsection{Strength of Evidence: Multiple-Comparison and Power}
\label{app:strength_of_evidence}

Our findings span a wide range of statistical power, and we interpret them accordingly.
The \textit{population-level null}---existing-benchmark QA accuracy and matched-model \textsc{MemUse} scores do not predict user satisfaction---is well powered ($N = 1{,}270$ sessions across 29 users, equivalence-confirmed; \S\ref{sec:result1_null}).
The \textit{per-instance retrieval--integration decoupling} ($\rho = -0.009$ on 207 paired questions nested in 48 sessions; \S\ref{sec:gap_perinstance}) is likewise well powered and robust: the same model reading the same reconstructed context answers 79\% of questions when queried directly but references only 8\% when each question is evaluated on its natural conversational response.
These two findings form the empirical core of the paper and survive Bonferroni correction at $\alpha / 6 = 0.0083$ (a six-test family-wise threshold), as does the engagement-covariate model comparison reported in Appendix~\ref{app:engagement_full} ($p < .001$).

The \textit{within-instance integration-to-satisfaction association} (\S\ref{sec:integration_effect}, $N = 48$ memory-moment sessions, 16 users) is a lower-power observational test: LMM $\beta = +0.556$, $p = .082$, 95\% Wald CI roughly $[-0.07,\, +1.19]$, straddling zero; the non-parametric Spearman analogue ($p = .046$) does not survive Bonferroni correction at $\alpha/6$.
A user-clustered nonparametric bootstrap (10{,}000 iterations) gives a tighter percentile CI of $[+0.12, +0.98]$ that excludes zero (one-sided $p_{\text{boot}}=.004$); a within-user permutation is more conservative ($p=.23$, with only 11~informative clusters of 16~users).
We offer the finding as conditional evidence consistent with the mode-contrast reading rather than a standalone causal claim: the bootstrap excludes zero and the relationship is robust to user clustering, but the asymptotic Wald and the permutation analogue both indicate the evidence is moderate rather than strong, consistent with $N=48$.
The point estimate is robust to analyst choice ($\beta \in [+0.543, +0.563]$ across four session-level aggregation rules; Appendix~\ref{app:nat_aggregation}).

\paragraph{Causal interpretation.}
The 7-condition randomization supports causal claims about the null direct effect at the population level.
The matched-model integration-to-satisfaction association (\S\ref{sec:integration_effect}) and the engagement-dimension analyses involve non-randomized variables and should be interpreted as observational associations, not established causal pathways.

\paragraph{Post-hoc power.}
The study achieves 84.5\% power for $d = 0.25$ but only 20.9\% for $d = 0.10$; detecting the latter would require $\sim$70 users.

\section{Retrieval--Integration Gap: Diagnostics}
\label{app:gap_diagnostics}

\subsection{Failure Mode Analysis}
\label{app:failures}

\begin{table}[!t]
\centering
\small
\begin{tabular}{lcc}
\toprule
\textbf{Failure Mode} & \textbf{N} & \textbf{\%} \\
\midrule
Generic (ignores context) & 28 & 38.9 \\
Partial recall (gist only) & 19 & 26.4 \\
Hallucinated recall & 17 & 23.6 \\
Correct recall & 6 & 8.3 \\
Honest admission & 2 & 2.8 \\
\bottomrule
\end{tabular}
\caption{LC-100\% failure modes (GPT-4.1-mini, $N=72$). Despite having all prior turns available, 39\% of responses are generic and 24\% hallucinate recall.}
\end{table}

\section{Cross-Model and Prompt Ablations}
\label{app:cross_model_ablations}

\subsection{Cross-Model \textsc{MemUse}: Figure and Full Numbers}
\label{app:cross_model}

Figure~\ref{fig:cross_model} plots Direct QA and Natural Integration across the 7 conditions for three models; Table~\ref{tab:cross_model} reports the full numbers, including \textsc{MemUse} \textit{Reference}---per-question whether the fact appears in the conversational response.

\paragraph{N=73 vs.\ released N=72.}
The cross-model and prompt-ablation analyses (this appendix and \ref{app:ablations}) were conducted on the full set of 73 detected reactive instances. The released benchmark of 72 (\S\ref{sec:benchmark}, Table~\ref{tab:data_flow}) excludes one probe (annotator\_40, session 13) flagged in our post-hoc rating-reliability audit for likely AI-paste contamination on the user side---exactly the construct the benchmark is built to avoid. This single-instance difference shifts any per-condition Natural Integration cell by at most $1/72 \approx 1.4$~pp, leaves all reported conclusions and significance patterns unchanged, and preserves the 71-point Direct~QA vs.\ Reference gap.

\begin{table}[!t]
\centering
\small
\setlength{\tabcolsep}{4pt}
\begin{tabular}{llccc}
\toprule
\textbf{Model} & \textbf{Cond.} & \textbf{Direct} & \textbf{Ref.} & \textbf{NI} \\
 & & \textbf{(\%)} & \textbf{(\%)} & \textbf{(\%)} \\
\midrule
\multirow{7}{*}{GPT-4.1-mini}
 & Summary & 44.9 &  7.9 & 23.6 \\
 & LC-10   & 53.0 &  5.6 & 25.0 \\
 & LC-50   & 72.4 &  7.2 & 25.0 \\
 & LC-100  & 78.8 &  7.9 & 22.2 \\
 & RAG-10  & 49.2 &  7.0 & 27.8 \\
 & RAG-50  & 57.7 &  8.0 & 25.0 \\
 & RAG-100 & 62.2 &  7.3 & 26.4 \\
\midrule
\multirow{7}{*}{GPT-5.5}
 & Summary & 50.0 & 15.4 & 53.4 \\
 & LC-10   & 56.6 & 13.0 & 53.4 \\
 & LC-50   & 77.8 & 19.0 & 57.5 \\
 & LC-100  & 82.9 & 13.6 & 52.1 \\
 & RAG-10  & 54.4 & 13.4 & 49.3 \\
 & RAG-50  & 63.2 & 15.2 & 54.8 \\
 & RAG-100 & 64.5 & 12.9 & 54.8 \\
\midrule
\multirow{7}{*}{Gemini 3.1 Pro}
 & Summary & 37.9 &  8.2 & 32.9 \\
 & LC-10   & 52.3 &  9.2 & 31.5 \\
 & LC-50   & 67.7 &  8.8 & 30.1 \\
 & LC-100  & 72.3 &  7.6 & 37.0 \\
 & RAG-10  & 45.3 &  7.7 & 31.5 \\
 & RAG-50  & 55.1 & 10.1 & 37.0 \\
 & RAG-100 & 57.1 & 10.1 & 32.9 \\
\bottomrule
\end{tabular}
\caption{Cross-model \textsc{MemUse} accuracy across all 7 memory conditions, $N = 73$ instances each. \textbf{Direct}: \textsc{MemUse} Direct QA (per-question accuracy). \textbf{Ref.}: \textsc{MemUse} Reference (per question, against the natural response). \textbf{NI}: \textsc{MemUse} Natural Integration (holistic on the natural response, primary metric). GPT-4.1-mini values are from Table~\ref{tab:main_combined}; GPT-5.5 (\texttt{gpt-5.5-2026-04-23}) and Gemini~3.1~Pro (\texttt{gemini-3.1-pro-preview}) were re-run for this comparison with judge held fixed at GPT-5.4-nano. Direct QA spans 32.9--34.4 points across capacity for every model; NI spans at most 8.2 points within model and never tracks capacity.}
\label{tab:cross_model}
\end{table}

The cross-model evaluation uses the identical pipeline as the main paper (same 73 ground-truth instances, same context-construction code per condition, same judge prompt and judge model). Only the generation model varies. GPT-5.5 and Gemini~3.1~Pro are reasoning models; we allowed \texttt{max\_completion\_tokens}\,=\,4{,}000 to accommodate internal reasoning while leaving room for short conversational responses. Reproduction scripts and per-instance outputs are included in the released analysis code.

\subsection{Modern Memory Systems on \textsc{MemUse}}
\label{app:memory_systems}

\begin{table}[!t]
\centering
\small
\setlength{\tabcolsep}{4pt}
\begin{tabular}{@{}lccc@{}}
\toprule
\textbf{Memory system} & \textbf{Direct QA} & \textbf{Ref.} & \textbf{NI} \\
\midrule
Summary/LC/RAG (7 cond., range) & 44.9--78.8 & 5.6--8.0 & 22.2--27.8 \\
Mem0 & 41.5 & 7.3 & 58.3 \\
Letta (MemGPT) & 61.4 & 10.8 & 56.9 \\
\bottomrule
\end{tabular}
\caption{Modern memory systems on \textsc{MemUse} (72 instances, 316 questions; GPT-4.1-mini generator; judge fixed at GPT-5.4-nano; all values \%). Both raise Natural Integration (NI) well above the provisioning baselines, yet the Direct~QA vs.\ Reference gap persists (\S\ref{sec:gap_memsys}).}
\label{tab:memory_systems}
\end{table}

Table~\ref{tab:memory_systems} (\S\ref{sec:gap_memsys}) reports Mem0 and Letta on the 72 released instances, scored with the same release-format judge pipeline (GPT-5.4-nano, temp$=$0) and the same 316 fact targets as the paper.
\textbf{Mem0} (\texttt{mem0ai}~2.0.11, open-source \texttt{Memory} with a local Qdrant store; \texttt{text-embedding-3-small}; GPT-4.1-mini as Mem0's extraction LLM): prior sessions are ingested incrementally per user; on the trigger turn we retrieve once (top-10 memories) and reuse that memory context, under the verbatim deployed persona prompt, for both the natural reply and every Direct~QA answer---the same ``identical reconstructed context'' protocol as the paper.
\textbf{Letta} (\texttt{letta}~0.16.8, self-hosted server; \texttt{memgpt\_agent} agent type with the model pinned to GPT-4.1-mini and \texttt{text-embedding-3-small}): a fresh agent per instance loads the prior sessions into archival memory as passages; tools are restricted to \texttt{archival\_memory\_search} and \texttt{send\_message} so the agent actually consults archival memory; the natural reply is the agent's response to the trigger, and each Direct~QA question is asked as a separate explicit probe.
Because Letta generates through its own agent loop (with its own system prompt and tool calls), its generation is not perfectly matched to the deployed pipeline; Mem0 is the cleaner like-for-like comparison.
Exact configurations and per-instance outputs are released with the analysis code.

\subsection{Prompt and Architectural Ablations}
\label{app:ablations}

\begin{figure*}[!t]
\centering
\includegraphics[width=\textwidth]{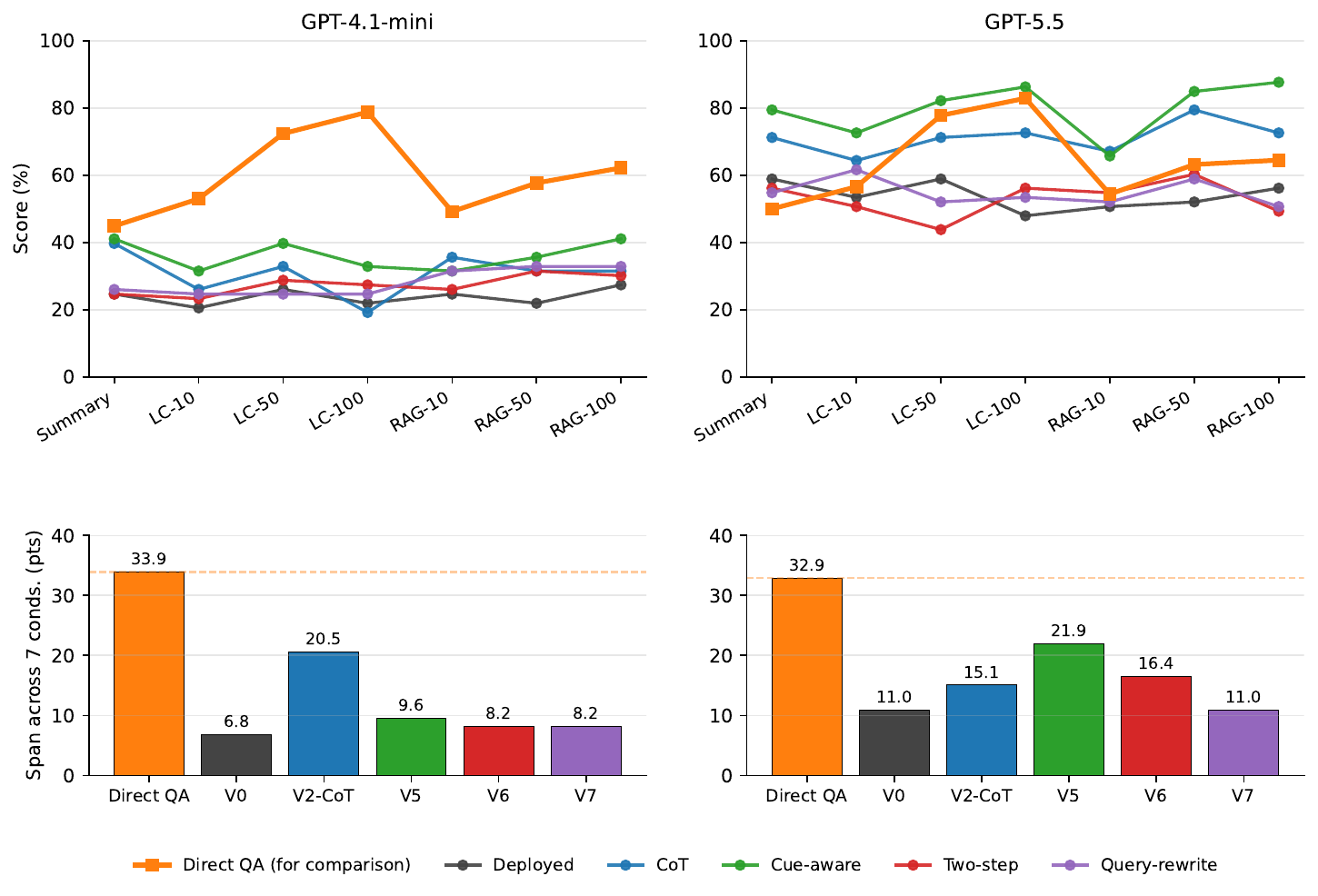}
\caption{Prompt interventions do not induce capacity-sensitivity in the integration channel. \textbf{Top:} Natural Integration (\%) by memory condition for V0 (deployed) and four interventions, with Direct~QA on the same axis as a retrieval baseline. \textbf{Bottom:} span across the 7 conditions (max$-$min, pts). Direct~QA spans 32.9--33.9~pts in both models; every integration variant spans at most 21.9~pts, and the V0 deployed pipeline spans only 6.8--11.0~pts. Variants: \textbf{V0}~deployed pipeline (no extra prompt); \textbf{V2-CoT}~chain-of-thought reasoning prepended; \textbf{V5}~one-line cue-aware patch; \textbf{V6}~two-step extract-then-integrate (a separate call extracts named details from context, then the deployed pipeline generates with those details handed in); \textbf{V7}~query rewrite~+~targeted RAG (the user trigger is rewritten into explicit search queries before retrieval). Per-condition numbers in \S\ref{app:prompt_variants}.}
\label{fig:prompt_variants_capacity}
\end{figure*}

\S\ref{sec:gap_ruledout} reports that prompt-level and retrieval-side interventions fail to close the retrieval--integration gap. This subsection gives the cross-condition figure (Figure~\ref{fig:prompt_variants_capacity}), the difficulty continuum across event types, the verbatim prompt texts, per-condition tables, and decomposition diagnostics. All variants use the same 73 reactive MemUse instances, the same reconstructed contexts per condition, and the same Natural Integration / Reference judge (GPT-5.4-nano, temp$=$0).

\subsubsection{Cross-Variant Overview}
\label{app:ablations_overview}

The 71-point Direct~QA / Natural Integration gap conceals a difficulty continuum: Direct~QA (78.8\%) $\to$ user memory probe ``Do you remember X?'' (51\%, $n{=}10$) $\to$ user re-provision ``As I mentioned, X'' (21\%, $n{=}62$; OR$=$4.06, $p{=}.001$ in a user-random-effect logistic GLM). Probes share the explicit-elicitation \textit{form} of Direct~QA but require integration into a conversational response, locating them midway on the continuum (full per-event-type tables in \S\ref{app:event_split}). Variants V2-CoT, V5, V6, V7 are defined and evaluated in \S\S\ref{app:prompt_variants}--\ref{app:v7_qr_rag}; the headline span diagnostic (Table~\ref{tab:nat_span_summary}) shows that across both models, every integration variant spans at most 21.9~pts versus the $\sim$33-pt Direct~QA span---some interventions raise the \textit{level} of Natural Integration (V5 on GPT-5.5 reaches the high 70s--80s) but the rise is flat across capacity, not capacity-driven. Two additional nuances: V5's lift is uniform across all 7 conditions including Summary-only where no prior facts are in context (judge fragility, \S\ref{app:judge_fragility}); and V7's rewritten queries retrieve \textit{worse} than raw triggers because in a longitudinal personal corpus user phrasings are already lexically aligned with prior turns (\S\ref{app:v7_qr_rag}). Prompt-level fixes (V2-CoT, failure-mode, few-shot, cue-aware) add $\sim$$+30$~pts to probe NI but at most $+8$~pts to re-provision NI at LC-100\%; the V6 two-step shows the inverse pattern. Prompt-level instruction amplifies semi-explicit recall it could already partly do; closing the implicit-integration gap requires architectural intervention.

\subsubsection{Deployed Persona (V0 Baseline)}
\label{app:persona_prompt}

The deployed system prompt (verbatim, used for every condition during the 4-month study and reused as V0 control in every ablation):

\begin{quote}\small\itshape
You are Luke, a friendly, kind, and understanding diary AI chatbot. Your goal is to make conversations feel like chatting with a friend. You are also a reflective companion who helps the user think about their day.\\
- Ask questions only a third of the time.\\
- Adapt your tone based on the user's mood.\\
- Aim for one or two sentences, like a real conversation.\\
- You are also here to act as a personal diary.
\end{quote}

A condition-specific framing is appended (``This is the $k$-th conversation. Summary of previous conversations: $\dots$. Generate the response using the following conversation history if necessary.''), followed by the reconstructed prior-turn context per condition (Summary-only, LC-$k\%$, RAG-$k\%$). All ablation variants below preserve this baseline and append additional instructions before the framing.

\subsubsection{Per-Variant Capacity-Span Summary (Both Models)}
\label{app:ruled_out_full}

Table~\ref{tab:nat_span_summary} reports the headline diagnostic of the V1--V7 prompt-variant ablation across all 7 memory conditions on both generation models: the Natural-Integration-rate span across the 7 capacity conditions remains well below the $\sim$33-pt Direct~QA span for every variant on either model, confirming that prompt-level interventions do not make the integration channel capacity-sensitive. V0 baselines are canonicalized per (model, condition) cell to a single deployment+judging run (the V5 cross-condition study for GPT-4.1-mini; the V2-CoT@GPT-5.5 followup for GPT-5.5), eliminating the $\le 2$-instance V0 drift that arose from independent per-variant judge re-runs (GPT-5.4-nano at temp$=$0 is not bit-exact across reruns). Each variant's $\Delta$ and McNemar $p$ are computed against this canonical V0 paired by instance. Full per-cell results (V0 NI, variant NI, $\Delta$ NI, McNemar $p$, V0 Reference, variant Reference for all $7 \times 7 \times 2 = 98$ cells) are released with the analysis code.

\begin{table}[!t]
\centering
\small
\setlength{\tabcolsep}{4pt}
\begin{tabular}{@{}lcc@{}}
\toprule
\textbf{Variant} & \textbf{GPT-4.1-mini} & \textbf{GPT-5.5} \\
\midrule
V0 (deployed)              &  6.8  & 13.7 \\
\midrule
V1                        &  9.6 &  8.2 \\
V2-CoT                    & 20.5 & 15.1 \\
V3                        &  9.6 & 13.7 \\
V4                        & 19.2 & 11.0 \\
V5                        &  9.6 & 21.9 \\
V6                        &  8.2 & 16.4 \\
V7                        &  8.2 & 11.0 \\
\midrule
\textit{Direct QA capacity span}$^{\dagger}$ & \textit{33.9} & \textit{32.9} \\
\bottomrule
\end{tabular}
\caption{\textbf{Natural Integration-span diagnostic (capacity insensitivity).} Each cell is max$-$min Natural Integration~(\%) across the 7 memory conditions for the indicated variant on the indicated model. $^{\dagger}$Same-model \textsc{MemUse} Direct~QA capacity span across the 7 conditions, reported in \S\ref{sec:gap_crossmodel}: 32.9~/~33.9~/~34.4~pts for GPT-5.5~/~GPT-4.1-mini~/~Gemini~3.1~Pro. \textit{Across every variant on either model, NI span remains $\le 22$~pts}---well below the $\sim$33-pt Direct QA span---confirming that prompt-level interventions do not make the integration channel capacity-sensitive in the way the retrieval channel is. Full per-cell numbers (V0/variant NI, $\Delta$, McNemar $p$) are released with the analysis code.}
\label{tab:nat_span_summary}
\end{table}

\subsubsection{Single-Call Prompt Variants (V1--V5)}
\label{app:prompt_variants}
\label{app:cue_aware}

V1--V4 are appended to the deployed persona before the framing, on LC-100\% only (the strongest existing condition); V5 is a one-line cue-aware patch evaluated across all 7 conditions. Verbatim text:

\begin{quote}\small\itshape
\textbf{V1 (light).} When the user references prior conversations or topics, integrate specific named details from your context (proper nouns, places, prior recommendations). Avoid generic empathetic responses. Do not invent details that are not in your context.\\[2pt]
\textbf{V2-CoT.} Before responding, follow these steps internally: (1) check if the user is referencing prior conversations; (2) if so, locate the relevant content in your context; (3) weave specific named details from that content into a natural conversational response; (4) if you cannot find the referenced content, acknowledge that rather than inventing details. Output only the final natural response.\\[2pt]
\textbf{V3 (failure-mode).} Three failure modes to avoid: (a) generic empathetic responses that ignore the user's reference to prior conversations; (b) inventing details not in context; (c) referring vaguely without naming specifics. When the user mentions prior topics, weave concrete named details from your context into a natural response.\\[2pt]
\textbf{V4 (few-shot).} Two integration exemplars prepended (Vienna/Figlmüller and Sapporo/Hakodate cases from real deployment, also used in Table~\ref{tab:asymmetry_examples}), followed by the instruction to apply the same pattern. The full prompt is included in the released analysis code.\\[2pt]
\textbf{V5 (cue-aware).} If the user refers to something from an earlier conversation, identify the relevant prior topic and respond with one or two specific remembered details. If unsure, say so.
\end{quote}

\paragraph{V1--V4 results.}
On LC-100\% with GPT-4.1-mini ($N=73$), Natural Integration: V0~24.7\%, V1~28.8\%, V2-CoT~34.2\%, V3~31.5\%, V4~27.4\%. McNemar exact two-sided $p$ vs.\ V0: 0.51, 0.12, 0.23, 0.77 respectively. None significant after Bonferroni at $\alpha/4=.0125$. Reference rates barely move ($+0.6$ to $+1.9$~pts). The V5 cue-aware variant is broken out across all 7 conditions in Table~\ref{tab:cue_aware_cross_condition}.

\begin{table}[!t]
\centering\small
\setlength{\tabcolsep}{4pt}
\begin{tabular}{lcccc}
\toprule
\textbf{Condition} & \textbf{V0} & \textbf{V5} & \textbf{$\Delta$} & \textbf{McNemar $p$} \\
\midrule
Summary-only & 24.7 & 41.1 & $+16.4$ & .004 \\
LC-10\%      & 20.5 & 31.5 & $+11.0$ & .096 \\
LC-50\%      & 26.0 & 39.7 & $+13.7$ & .053 \\
LC-100\%     & 21.9 & 32.9 & $+11.0$ & .039 \\
RAG-10\%     & 24.7 & 31.5 & $+6.8$  & .302 \\
RAG-50\%     & 21.9 & 35.6 & $+13.7$ & .031 \\
RAG-100\%    & 27.4 & 41.1 & $+13.7$ & .041 \\
\bottomrule
\end{tabular}
\caption{V5 cue-aware prompt vs.\ V0 deployed persona on Natural Integration~\% (GPT-4.1-mini, $N=73$ instances per cell). Variant~$\times$~condition interaction $\chi^2 = 2.5$, $df=6$, $p=.86$ (homogeneity of discordant-pair distribution). Reference rates barely move ($+0$ to $+5$~pts).}
\label{tab:cue_aware_cross_condition}
\end{table}

\subsubsection{Two-Step Extract-then-Integrate Pipeline (V6)}
\label{app:two_step}

V6 splits the task into two GPT-4.1-mini calls. Step~1 extracts named facts from the reconstructed context; Step~2 generates the response with those facts piped in as ``Notes from your prior conversations.''

\paragraph{Pipeline.}
\textit{Step~1 (extraction)}: a separate GPT-4.1-mini call (system: \textit{``You extract relevant facts from prior conversation context.''}) is shown the reconstructed context and the user trigger and asked to ``list up to 5 specific named details from the context that relate to what the user is referencing or might want acknowledged'' (proper nouns, places, prior recommendations, dates, activities) as a bullet list, or output \texttt{NO\_RELEVANT\_CONTEXT} if nothing relates.
\textit{Step~2 (generation)}: the deployed persona prompt (Appendix~\ref{app:persona_prompt}) is augmented with a \texttt{[Notes from your prior conversations]} block containing the Step-1 output (replaced by \textit{``(none relevant)''} if \texttt{NO\_RELEVANT\_CONTEXT}), followed by the current conversation. Per-condition Natural Integration and Reference for V6 vs.\ V0 are reported in Table~\ref{tab:v6_two_step}.

\begin{table}[!t]
\centering\small
\setlength{\tabcolsep}{3pt}
\begin{tabular}{lccccc}
\toprule
\textbf{Cond.} & \textbf{V0~NI} & \textbf{V6~NI} & \textbf{$\Delta$~NI} & \textbf{V6~Ref.} & \textbf{McN.~$p$} \\
\midrule
Summary-only & 24.7 & 24.7 & $+0.0$ & 10.0 & 1.00 \\
LC-10\%      & 20.5 & 23.3 & $+2.7$ & 6.5  & .77  \\
LC-50\%      & 26.0 & 28.8 & $+2.7$ & 9.0  & .82  \\
LC-100\%     & 21.9 & 27.4 & $+5.5$ & 8.1  & .29  \\
RAG-10\%     & 26.0 & 26.0 & $+0.0$ & 7.5  & 1.00 \\
RAG-50\%     & 21.9 & 31.5 & $+9.6$ & 8.7  & .09  \\
RAG-100\%    & 26.0 & 30.1 & $+4.1$ & 8.4  & .63  \\
\bottomrule
\end{tabular}
\caption{V6 two-step pipeline vs.\ V0 deployed persona (GPT-4.1-mini, $N=73$). Natural Integration and Reference in \%. V6 is comparable to or worse than V2-CoT ($34.2\%$ on LC-100\%; Appendix~\ref{app:prompt_variants}).}
\label{tab:v6_two_step}
\end{table}

\paragraph{Extraction-quality decomposition (LC-100\%).}
We score each Step-1 extraction against the instance's ground-truth facts and cross with whether Step-2 referenced those facts in its natural response (Reference $=$ yes/no; Table~\ref{tab:v6_decomposition}):

\begin{table}[!t]
\centering\small
\setlength{\tabcolsep}{3pt}
\begin{tabular}{lcc}
\toprule
\textbf{Step-1 found GT?} & \textbf{Step-2 referenced?} & \textbf{Count} \\
\midrule
Yes & Yes & 11 \\
\textbf{Yes} & \textbf{No}  & \textbf{37} \\
No  & Yes & 9  \\
No  & No  & 16 \\
\textsc{No\_relevant\_context} & --- & 2 \\
\bottomrule
\end{tabular}
\caption{V6 extraction-quality decomposition on LC-100\% ($N=73$). Step-1 mean GT-coverage $=0.51$. When Step-1 successfully named the ground-truth facts, Step-2 still failed to integrate them in 37/48 = 77\% of cases.}
\label{tab:v6_decomposition}
\end{table}

The 77\% drop rate is the cleanest mechanistic localization we obtain: the model has been pre-handed exactly the named details it needs; generation refuses to use them.

\subsubsection{Query-Rewrite + Targeted Retrieval (V7)}
\label{app:v7_qr_rag}

V7 tests whether the deployed RAG conditions underperform because raw user-trigger embeddings are bad retrieval queries. We replace the deployed retrieval query with multi-aspect rewrites generated by GPT-4.1-mini.

\paragraph{Pipeline.}
\textit{Step~1 (query rewriting)}: a separate GPT-4.1-mini call (system: \textit{``You generate memory-retrieval queries from a user's conversational message.''}) is shown the user trigger and asked, if it implicitly references prior conversations, to generate 1--3 explicit self-contained search queries suitable for embedding-based retrieval (one per line; output \texttt{NO\_QUERIES} if no reference).
\textit{Step~2 (retrieval and generation)}: each rewritten query is embedded with \texttt{text-embedding-3-small} (deployed model) and used to retrieve top-5 from the per-user importance-thresholded prior-turn pool, dedupe by (session, turn) up to 10 unique passages. The retrieved passages are prepended to the persona prompt as a \texttt{[Notes from your prior conversations, retrieved by your memory system]} block before generation. For LC-$k\%$ cells, V7 \textit{also} retains the full LC-$k\%$ context dump (V7-LC tests rewrite-notes-on-top-of-long-context); V7-RAG-$k\%$ replaces the deployed RAG retrieval query only. Per-condition Natural Integration and retrieval recall are reported in Table~\ref{tab:v7_qr_rag}.

\begin{table}[!t]
\centering\small
\setlength{\tabcolsep}{4pt}
\begin{tabular}{lcccc}
\toprule
 & \multicolumn{2}{c}{\textbf{NI (\%)}} & \multicolumn{2}{c}{\textbf{Recall (\%)}} \\
\cmidrule(lr){2-3}\cmidrule(lr){4-5}
\textbf{Cond.} & \textbf{V0} & \textbf{V7} & \textbf{V7} & \textbf{V0} \\
\midrule
Summary-only & 24.7 & 26.0 &  8.5 &  9.0 \\
LC-10\%      & 26.0 & 24.7 & 29.9 & 32.7 \\
LC-50\%      & 26.0 & 24.7 & 70.1 & 75.4 \\
LC-100\%     & 23.3 & 24.7 & 67.3 & 78.7 \\
RAG-10\%     & 27.4 & 31.5 & 29.9 & 32.7 \\
RAG-50\%     & 24.7 & 32.9 & 70.1 & 75.4 \\
RAG-100\%    & 26.0 & 32.9 & 67.3 & 78.7 \\
\bottomrule
\end{tabular}
\caption{V7 query-rewrite RAG vs.\ V0 (GPT-4.1-mini, $N=73$). Natural Integration (NI) and retrieval recall (token-overlap $@0.5$ between retrieved passages and ground-truth fact spans), both in \%. V7 lifts NI modestly on RAG-50/100\% ($+8.2/+6.8$~pts; McNemar n.s.) but rewritten queries actually \textit{retrieve worse} than raw user triggers at high $k\%$ (LC/RAG-100\% recall $67.3$ vs.\ $78.7$).}
\label{tab:v7_qr_rag}
\end{table}

\paragraph{Why query rewriting hurts here.}
Standard RAG benchmarks assume a corpus where naive retrieval is bad: short user questions over Wikipedia-style declarative passages, where rewriting (e.g., HyDE) helps because hypothetical answers embed closer to passage content than questions do. In a longitudinal personal corpus the situation reverses. The user's trigger (``\textit{the music Luke recommended the other day}'') already contains the lexical anchors of the relevant prior turn (\textit{music}, \textit{recommend}). Rewriting into question form (``\textit{What music did Luke recommend?}'') replaces a conversational match with a Q-form query against an A-form corpus, while spreading the top-10 retrieval budget across 1--3 distinct queries that may individually rank the relevant passage lower than the raw trigger does. The generic-empathetic failure mode persists at $\sim$48\%, identical to V0, regardless of whether retrieval recall improves.

\subsubsection{GPT-5.5 + V2-CoT}
\label{app:gpt55_v2cot}

To rule out a model-capacity ceiling for the prompt-ablation null, we re-ran V0 vs.\ V2-CoT (Appendix~\ref{app:prompt_variants}) on GPT-5.5 across all 7 conditions (Table~\ref{tab:gpt55_v2cot}). V0 generations are reused from the cross-model run (Appendix~\ref{app:cross_model}); V2-CoT generations are fresh; all judgments are re-run for apples-to-apples comparison.

\begin{table}[!t]
\centering\small
\setlength{\tabcolsep}{3pt}
\begin{tabular}{lcccc}
\toprule
\textbf{Cond.} & \textbf{V0~NI} & \textbf{V2-CoT~NI} & \textbf{$\Delta$} & \textbf{McN.~$p$} \\
\midrule
Summary-only & 54.8 & 71.2 & $+16.4$ & .023 \\
LC-10\%      & 47.9 & 64.4 & $+16.4$ & .023 \\
LC-50\%      & 61.6 & 71.2 & $+9.6$  & .19  \\
\textbf{LC-100\%}     & \textbf{47.9} & \textbf{72.6} & \textbf{$+24.7$} & \textbf{.0005} \\
RAG-10\%     & 50.7 & 67.1 & $+16.4$ & .023 \\
RAG-50\%     & 61.6 & 79.5 & $+17.8$ & .011 \\
RAG-100\%    & 58.9 & 72.6 & $+13.7$ & .031 \\
\bottomrule
\end{tabular}
\caption{V2-CoT on GPT-5.5 across all 7 conditions ($N=73$). On LC-100\%, $+24.7$~pt lift closes $\sim$70\% of the V0 $\to$ Direct~QA gap (47.9 $\to$ 72.6 vs.\ 83.0 ceiling), leaving $+10.3$~pts of headroom. Variant~$\times$~condition interaction $\chi^2 = 2.2$, $df=6$, $p=.90$---lift is uniform across conditions including Summary-only where no facts are in context, replicating the cue-aware Summary-fragility pattern (Appendix~\ref{app:judge_fragility}). Reference remains low (LC-100\% $24.6\%$) confirming the dissociation between holistic Natural Integration and item-level Reference seen on GPT-4.1-mini.}
\label{tab:gpt55_v2cot}
\end{table}

\subsubsection{Holistic Judge Fragility on Recall-Style Prompts}
\label{app:judge_fragility}

V5 (cue-aware) and V2-CoT both lift Natural Integration on \textit{Summary-only} by $+16.4$~pts even though Summary-only contains no prior-turn facts to integrate, with no corresponding rise in fact-based hallucination ($\Delta=+0$ in failure-mode classification). The lift magnitude is statistically indistinguishable from the lift on context-bearing cells (variant~$\times$~condition interaction $p=.86$ for V5, $p=.90$ for V2-CoT on GPT-5.5). The holistic Natural Integration judge---a single binary read of whether the response demonstrates memory of the topic---appears partly responsive to recall-style framing language (``As you mentioned\dots'', ``I remember you said\dots'') without verifying substantive integration. This is consistent with the moderate inter-judge agreement reported in Appendix~\ref{app:judge} (Fleiss $\kappa=.26$ on holistic Natural Integration vs.\ $.65$ on per-question Reference). The Summary-fragility lift is therefore best read as judge fragility plus a smaller real component; the LC-100\%-specific $+24.7$~pt lift on GPT-5.5 is suggestive of a real capacity-conditional gain on top of fragility, but the asymptotic ceiling at $\sim$73\% remains $\sim$10~pts below the $83\%$ Direct~QA ceiling on the same instances.

\subsubsection{Event-Type Split (Probe vs.\ Re-provision)}
\label{app:event_split}

The 73 reactive instances comprise 11 \textit{user memory probes} (``Do you remember X?'') and 62 \textit{user re-provisions} (``As I mentioned, X''). The deployed V0 baseline already shows the predicted continuum: probe Natural Integration $\sim$50\% vs.\ re-provision Natural Integration $\sim$20\% (logistic GLM with user random effect: $\text{OR}=4.06$, 95\% CI $[1.71, 9.63]$, $p=.001$). Table~\ref{tab:event_split} reports the LC-100\% breakdown across all prompt variants.

\begin{table}[!t]
\centering\small
\setlength{\tabcolsep}{4pt}
\begin{tabular}{lccc}
\toprule
\textbf{Variant (LC-100\%)} & \textbf{Overall} & \textbf{Probe} & \textbf{Re-prov} \\
\midrule
V0 baseline                   & 23.6 & 40.0 & 21.0 \\
V1 (light)                    & 27.8 & 40.0 & 25.8 \\
V2-CoT                        & 33.3 & 70.0 & 27.4 \\
V3 (failure-mode)             & 30.6 & 70.0 & 24.2 \\
V4 (few-shot)                 & 26.4 & 70.0 & 19.4 \\
V5 (cue-aware)                & 31.9 & 70.0 & 25.8 \\
V6 (two-step)                 & 26.4 & 50.0 & 22.6 \\
V7 (query-rewrite RAG)        & 23.6 & 50.0 & 19.4 \\
\bottomrule
\end{tabular}
\caption{Natural Integration~\% on LC-100\%, split by event type ($n_{\text{probe}}=11$, $n_{\text{re-prov}}=62$; full 73-instance detection set, see Appendix~\ref{app:cross_model}). Reasoning/framing prompts (V2-CoT, V3, V4, V5) lift probe Natural Integration by $\sim+30$~pts but re-provision by $\leq+8$~pts. The architectural V6 two-step shows the inverse pattern (mean across 7 conditions: $-5.7$~pts on probes, $+4.4$~pts on re-provisions). Probe $n=11$ implies $\pm$25--30~pp 95\% CIs per single cell; the pattern's reliability comes from directional consistency across multiple variants and 7 conditions.}
\label{tab:event_split}
\end{table}

The split clarifies what each intervention family does: prompt-level fixes amplify what the model could already partly do (semi-explicit ``Do you remember X?'' cues), without repairing the implicit-integration failure mode (``As I mentioned X''). Architectural retrieval-restructuring (V6) helps the harder implicit cases at small cost to the easier ones, but the absolute improvement remains small. Prompt-level instruction amplifies semi-explicit recall; closing the implicit-integration gap requires architectural intervention.

\section{Exploratory Engagement and Proactive Analyses}
\label{app:engagement}

The analyses in this appendix---predictors of satisfaction beyond memory condition, system proactive recall, and user re-provisioning---are \textbf{exploratory}: they involve non-randomized variables (response length, specificity, integration success, recall timing) and should be interpreted as observational associations, not causal claims.

\subsection{Engagement Predictors of Satisfaction}
\label{app:engagement_predictors}

\begin{figure}[!t]
\centering
\includegraphics[width=\columnwidth]{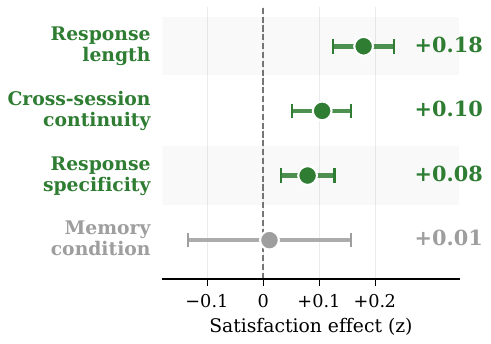}
\caption{Predictors of $z$-scored satisfaction ($N = 1{,}270$ sessions, 29 users). Bars show 95\% Wald CIs around point estimates. Three engagement metrics each independently predict higher satisfaction; memory condition does not.}
\label{fig:forest}
\end{figure}

The deployment-level null (\S\ref{sec:result1_null}) raises an obvious follow-up: \textit{what does} predict per-session satisfaction if memory condition does not? Fitting an LMM (Eq.~\ref{eq:lmm}) on all 1{,}270 rating-valid sessions with four $z$-standardized fixed-effect predictors---response length (\textsc{Len}), response specificity (\textsc{Spec}; system proper-noun rate, capturing whether responses reference user-specific personal details), cross-session continuity (\textsc{Cont}; lexical and entity overlap with prior sessions plus explicit user backward references), and memory condition (\textsc{MemCond})---three engagement predictors are independently significant (Figure~\ref{fig:forest}): response length ($\beta = 0.179$, $p < .001$), cross-session continuity ($\beta = 0.105$, $p < .001$), and specificity ($\beta = 0.079$, $p = .001$). Memory condition contributes nothing ($\beta = 0.011$).

\paragraph{Long-but-generic responses are the failure mode.}
Cross-session continuity interacts \textit{negatively} with response length ($\beta = -0.067$, $p = .002$): when users reference prior topics, long responses that fail to address those references reduce satisfaction. The negative continuity-$\times$-length effect is offset by specificity (three-way interaction $\beta_{\text{Cont}\times\text{Len}\times\text{Spec}} = 0.035$, $p = .084$; binary-condition specification $\beta = 0.073$, $p = .001$). ``Longer'' helps when paired with specificity; without it, length backfires precisely on the sessions where memory is at stake.

\subsection{Engagement Dimensions: Full LMM Specification and Operationalization}
\label{app:engagement_full}

This subsection supplements Appendix~\ref{app:engagement_predictors}. The full model is Eq.~\ref{eq:lmm} with fixed-effect predictors $\{\textsc{Len}, \textsc{Spec}, \textsc{Cont}, \textsc{MemCond}\}$. Variable definitions: \textsc{Len}$_{ij}$ is mean system response length (words) over the session; \textsc{Spec}$_{ij}$ is specificity, computed as the proper-noun rate in system responses; \textsc{Cont}$_{ij}$ is cross-session continuity, combining lexical overlap with prior sessions, entity overlap, and explicit user backward references (full operationalization in Appendix~\ref{app:operationalization}); \textsc{MemCond}$_{ij}$ is the memory condition assigned to session~$j$ for user~$i$.

\paragraph{Engagement vs.\ existing benchmarks (model comparison).}
We compare two LMMs fitted with maximum likelihood.
Model~$\mathcal{A}$ uses only condition-level QA benchmark accuracy: $y_{ij} \sim \beta\,\textsc{QA}_{ij} + (1 \mid \text{user}_i)$.
Model~$\mathcal{B}$ uses the three engagement dimensions $\{\textsc{Len}, \textsc{Spec}, \textsc{Cont}\}$ (without memory condition).
Model~$\mathcal{B}$ significantly outperforms Model~$\mathcal{A}$ ($\Delta$AIC $= 14.1$; likelihood ratio $\chi^2(2) = 18.14$, $p < .001$).
Adding QA accuracy on top of Model~$\mathcal{B}$ provides \textit{no additional predictive value} ($\chi^2(1) = 1.02$): QA accuracy captures none of the variance that engagement dimensions explain.

\paragraph{Specificity is not a hidden capacity effect.}
Specificity does not differ across memory conditions ($F = 0.43$), so its coefficient is not absorbing variance attributable to capacity. The engagement story is genuinely orthogonal to memory condition.

\subsection{Memory Integration and Engagement: Extended Analyses}
\label{app:amplifier}

\paragraph{Matched-model main effect details.}
Extending the base result reported in \S\ref{sec:integration_effect} ($\beta = +0.556$, $p = .082$), we examine sensitivity to engagement covariates.
Adding response length as a control attenuates the Natural Integration coefficient only slightly ($\beta = +0.574$, $p = .091$, length $\beta = +0.420$, $p = .028$), reflecting partial overlap between integration and substantive responding.
Adding both response specificity and prior-context dependence as controls further attenuates the Natural Integration coefficient to $\beta = +0.430$, $p = .121$, with prior-context dependence emerging as significant ($\beta = +0.236$, $p = .017$).
The pattern is consistent: integration is positively associated with satisfaction on memory-moment sessions, with the effect partially mediated by---but not eliminated by---engagement covariates.
Of the 48 sessions, 11 received Natural Integration = 1 (successful integration) and 37 received Natural Integration = 0; the small successful-integration cell limits power for stricter specifications.
Critically, the same exercise on Direct QA accuracy yields a null effect (Spearman $\rho = +0.033$; LMM $\beta = -0.093$), and the head-to-head LMM (\S\ref{sec:within_instance}) confirms that Direct QA contributes nothing once Natural Integration is included ($\beta_{\text{DQ}} = -0.353$).

\begin{figure}[!t]
\centering
\includegraphics[width=\columnwidth]{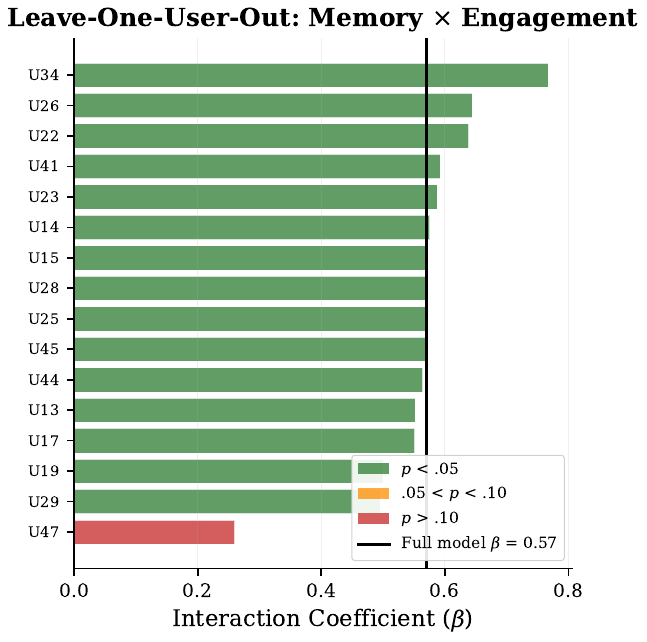}
\caption{Leave-one-user-out cross-validation of the memory $\times$ engagement interaction. Each bar shows the interaction coefficient when one user is withheld. The effect is positive in all 16 iterations and significant ($p < .05$) in 15 of 16 (green). The vertical line marks the full-model estimate ($\beta = 0.57$).}
\label{fig:louo}
\end{figure}

\begin{figure*}[!t]
\centering
\includegraphics[width=\textwidth]{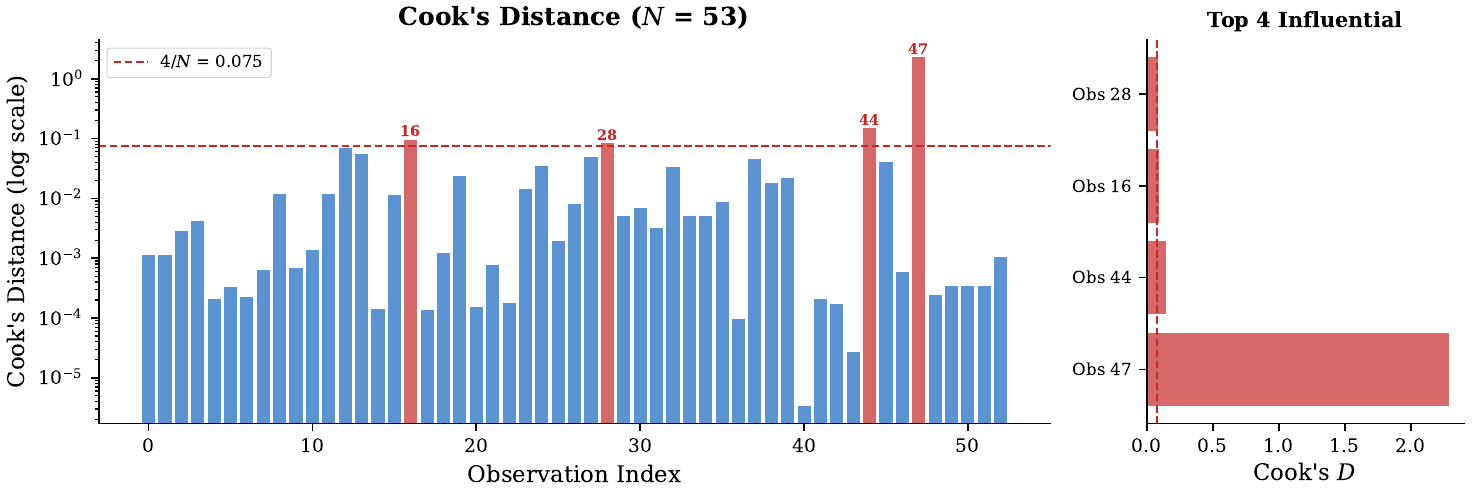}
\caption{Cook's distance for the memory $\times$ engagement model ($N = 53$). Four observations exceed the $4/N$ threshold (dashed line). After removing all four, the interaction remains significant ($\beta = 0.35$, $p = .018$).}
\label{fig:cooks}
\end{figure*}

\paragraph{Robustness: leave-one-user-out and influence diagnostics.}
The matched-model result (\S\ref{sec:integration_effect}) is not driven by individual users or outlier sessions.
Figure~\ref{fig:louo} shows leave-one-user-out cross-validation: the interaction coefficient is positive in all 16 iterations and significant ($p < .05$) in 15 of 16.
Figure~\ref{fig:cooks} plots Cook's distance on the $N = 53$ memory-moment fit (53 of the 55 filtered instances; 2 dropped for missing engagement covariates); four observations exceed the $4/N$ threshold, and after removing all four the interaction remains significant ($\beta = 0.35$, $p = .018$).

\paragraph{Cross-session continuity \texttimes\ length interaction.}
A separate length interaction characterizes how user-initiated reference to prior topics moderates the length--satisfaction relationship.
On all 1,270 sessions, cross-session continuity interacts \textit{negatively} with length ($\beta = -0.067$, $p = .002$): when users reference prior topics, long responses that fail to address those references are associated with \textit{reduced} satisfaction.
A three-way interaction resolves this apparent contradiction: the negative continuity $\times$ length relationship is offset by response specificity ($\beta_{\text{continuity} \times \text{length} \times \text{specificity}} = 0.035$, $p = .084$), confirmed with a stronger effect using an alternative specification with binary memory condition ($\beta = 0.073$, $p = .001$).
Long, \textit{specific} responses remain beneficial even in high-continuity sessions, while long, \textit{generic} responses that ignore the user's conversational history are detrimental.

\paragraph{Capability probe (GPT-5.4 generator).}
For completeness, we report a parallel analysis on offline GPT-5.4 generations on the same 53 (per-instance) memory-moment events.
Note that these outputs were never seen by users, so they cannot be tied directly to user experience and are reported here only as a capability characterization.
Under this counterfactual generator, the matched main effect of Natural Integration disappears ($\beta = -0.002$), but an NI $\times$ Length interaction emerges ($\beta = +0.570$, $p = .013$; simple slopes: length $\to$ rating $\rho = .45$, $p = .028$ when Natural Integration = 1 vs.\ $\rho = .15$ when Natural Integration = 0).
We interpret this as suggestive evidence that with stronger model capability, memory integration may compound with substantive responding rather than acting as an additive main effect; we do not include this finding in the main text because the underlying responses do not correspond to anything users actually rated.

\subsection{Prior-Context Dependence (PCD): Detector and Null Result}
\label{app:pcd}

To complement cross-session continuity (a user-side measure), we developed \textit{prior-context dependence} (PCD), a demand-side judge-scored measure (0--3) of how much a session requires prior conversational context. The detector is a GPT-5.4 judge selected by sweeping prompt structure (three variants) and prior-summary context size ($k \in \{1,3,5,10\}$) against 253 expert-annotated stratified sessions; the selected v2 prompt at $k{=}5$ achieves binary F1$=$.835, quadratic-weighted $\kappa = .711$, and 85.0\% binary accuracy. Scaled to the 1{,}251 filtered sessions, PCD distribution is 64.6\%/33.0\%/2.3\%/0.1\% across scores 0--3. \textbf{In an LMM with response length, specificity, and PCD, the PCD coefficient is null ($\beta = -0.011$): memory demand per se does not predict satisfaction.} An exploratory check on a separately judged \textit{integration quality} (IQ; 0--3) shows the same pattern as the matched-model NI effect (\S\ref{sec:integration_effect}): on the 30 sessions with PCD\,$\geq 2$, IQ$=$2 carries $+0.21$~$z$ over IQ$\leq 1$ (Spearman $\rho{=}.09$, underpowered at $N{=}30$), but no IQ effect emerges among the larger PCD$=$1 set. Full prompt-sweep numbers and IQ tables are released with the analysis code.

\subsection{Re-Provisioning Burden}
\label{app:reprov}

We find a significant negative interaction between user word count and cross-session continuity ($\beta = -.086$, $p < .001$).
In high-continuity sessions---where users reference prior topics---longer user utterances predict \textit{lower} satisfaction.
When users must re-explain previously shared information because the system fails to remember, the effort diminishes satisfaction.
This interaction is robust to controlling for number of turns ($\beta = -.076$, $p = .002$), user sentiment ($\beta = -.088$, $p < .001$), and all controls combined ($\beta = -.074$, $p = .004$); it also survives winsorization at the 1st/99th percentiles ($\beta = -.092$, $p < .001$).
This is corroborated by a marginal difference between user-initiated and system-initiated memory events ($p = .086$): user re-provisions (mean $z = +0.17$) are associated with higher satisfaction than system proactive recalls (mean $z = -0.14$).

\subsection{Proactive Recalls: Methods and Extended Analyses}
\label{app:proactive_full}

This appendix supplements the main-text proactive analysis (\S\ref{sec:proactive}) with verbatim judge prompts, per-condition breakdowns, the aggregate reactive--proactive contrast, robustness checks, and the canonical well-timed / ill-timed examples.
The 70 system-initiated proactive recalls are detected by the same event-detection pipeline as the reactive moments (\S\ref{sec:benchmark}); each is scored on grounding and appropriateness using GPT-5.4 at temperature~0.

\begin{figure*}[!t]
\centering
\includegraphics[width=\textwidth]{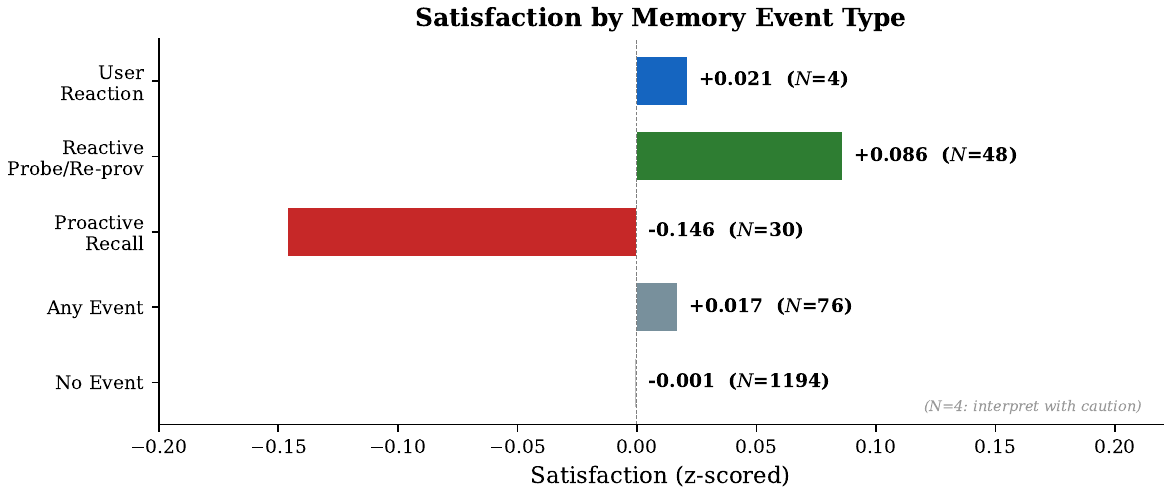}
\caption{Satisfaction by memory event type (session-level aggregate). Reactive events (user-initiated probes and re-provisions) are associated with higher satisfaction than proactive system recalls ($\bar{z} = +0.09$ vs.\ $-0.14$, $p = .086$ on sessions; \S\ref{sec:proactive} refines this by scoring the recalls themselves).}
\label{fig:proactive}
\end{figure*}

\paragraph{Aggregate reactive vs.\ proactive contrast.}
Figure~\ref{fig:proactive} shows the session-level aggregate underlying the main-text analysis: sessions containing reactive (user-initiated) memory events have higher $z$-scored satisfaction than sessions containing proactive system recalls.

\paragraph{Judge prompts (0/1/2).}
Both judges are GPT-5.4, temperature~0, with system prompts framing the rater as an expert annotator. \textit{Grounding} is shown the system's recall turn, the annotated referenced topic, and the prior-session summaries (newest last), and asked: 0~$=$~not grounded (hallucination, or too vague to be a factual claim); 1~$=$~partially grounded (general topic present but specifics wrong/missing); 2~$=$~fully grounded (specific content clearly present in prior summaries). \textit{Appropriateness} is shown the current session up to and including the recall turn, and asked: 0~$=$~non-sequitur/forced (derails or ignores what the user just said); 1~$=$~tangential (loosely related but not clearly invited); 2~$=$~natural/invited (fits the flow and is clearly relevant). Both prompts end with \textit{``Answer with ONLY a single digit.''} Verbatim text is in the released code.

\paragraph{Rating $z$ by score bin (event-level, $N{=}64$).}
Table~\ref{tab:proactive_bins} cross-tabulates mean satisfaction against the grounding and appropriateness scales.

\begin{table}[!t]
\centering\small
\setlength{\tabcolsep}{4pt}
\begin{tabular}{lccc|ccc}
\toprule
& \multicolumn{3}{c}{\textbf{Grounding}} & \multicolumn{3}{c}{\textbf{Appropriateness}} \\
\cmidrule(lr){2-4}\cmidrule(lr){5-7}
\textbf{Score} & 0 & 1 & 2 & 0 & 1 & 2 \\
\midrule
$N$ & 1 & 20 & 43 & 8 & 25 & 31 \\
$\bar z$ & $+0.09$ & $-0.14$ & $-0.08$ & $-0.48$ & $+0.03$ & $-0.10$ \\
\bottomrule
\end{tabular}
\caption{Mean $z$-scored satisfaction by Proactive-\textsc{MemUse} score bin (64~events, 24~users, 63~unique sessions with $\textsc{talking\_index} \le 6$). The only pronounced effect is the low-appropriateness bin; better scores on either axis do not translate into higher satisfaction.}
\label{tab:proactive_bins}
\end{table}

\paragraph{Proactive events by condition.}
Table~\ref{tab:proactive_by_cond} confirms that the null holds across every deployed memory condition.

\begin{table}[!t]
\centering\small
\setlength{\tabcolsep}{4pt}
\begin{tabular}{lcccc}
\toprule
\textbf{Condition} & $\boldsymbol{N}$ & \textbf{Ground.} & \textbf{Approp.} & $\boldsymbol{\bar{z}}$ \\
\midrule
Summary-only & 3 & 1.67 & 1.33 & $+0.14$ \\
LC-10\% & 21 & 1.48 & 1.14 & $-0.31$ \\
LC-50\% & 17 & 1.76 & 1.24 & $+0.12$ \\
LC-100\% & 9 & 1.78 & 1.33 & $-0.18$ \\
RAG-10\% & 4 & 1.75 & 2.00 & $+0.20$ \\
RAG-50\% & 6 & 1.67 & 1.67 & $-0.02$ \\
RAG-100\% & 4 & 1.75 & 2.00 & $-0.28$ \\
\bottomrule
\end{tabular}
\caption{Proactive-\textsc{MemUse} scores and mean $z$-scored satisfaction by deployed memory condition (event-level, $N{=}64$). Both axes are stable across conditions; RAG conditions achieve uniformly high appropriateness (mean $\approx 2.0$ on RAG-10/100) without a corresponding satisfaction gain.}
\label{tab:proactive_by_cond}
\end{table}

\paragraph{Detection summary.}
The 70 proactive recalls are predominantly grounded (46 fully, 21 partial, 3 hallucinated), but timing is more variable (32 invited, 29 tangential, 9 mistimed).

\paragraph{Asymmetric-landscape numerics.}
The three effects shown in Figure~\ref{fig:asymmetry} (main text): reactive integration success has the largest point estimate ($\beta=+0.56$) but its 95\% interval ($-0.07$, $+1.19$) and the mistimed-recall interval ($-1.17$, $+0.21$) both straddle zero, while the re-provisioning burden ($\beta=-0.086$, CI $[-0.126,-0.046]$) is the only effect whose interval excludes zero---the small but reliable penalty users incur when the system fails to remember.

\paragraph{Robustness of the proactive null.}
The reactive--proactive gap holds apples-to-apples with identical judge and prompt: scoring the 64 proactive recalls with the same GPT-5.4 Natural Integration metric used on reactive memory moments yields $\rho = -0.20$ ($p = .12$; 62 of 64 judged yes), against the reactive baseline $\rho = +0.29$ ($p = .046$)---the asymmetry is not an artifact of the 0/1/2 Proactive-\textsc{MemUse} scales.
A second judge (Gemini~3.1~Pro) replicates the null on both axes ($\rho_{\text{ground}} = -0.02$; $\rho_{\text{approp}} = +0.08$), and the only directional signal is again at appropriateness${=}0$ ($\bar{z} = -1.70$, $N{=}3$).
Inter-judge agreement is substantial on grounding (weighted $\kappa = 0.73$) and fair on appropriateness ($\kappa = 0.28$), as expected for a more subjective judgment; the rating null is robust across both.
TOST equivalence at $\delta = 0.25$ is underpowered at $N{=}64$ (all $p > .09$); we therefore read this as a directional null with a robust apples-to-apples asymmetry, not a formal equivalence claim.
On the joint LMM, neither grounding nor appropriateness survives conditioning on the other ($\beta_{\text{ground}} = +0.29$, $p = .12$; $\beta_{\text{approp}} = +0.16$).

\paragraph{Topical-continuation coding.}
To check whether users actually engage with the recalled topic, we code each recall's immediate next user turn as topically engaging if it shares any content word with the referenced topic. From the topic we extract alphabetic tokens of length $>4$ with a small stopword list (system/role and meta terms) removed; from the user turn we extract lowercased alphabetic tokens of length $\geq 3$. The two sets overlap in 19 of 62 recalls (31\%), and the binary topical-continuation indicator is uncorrelated with appropriateness ($\rho = -0.02$): users walk past well-timed callbacks at the same rate as poorly-timed ones.

\paragraph{What well-timed and ill-timed recalls look like.}
A well-timed (appropriateness${=}2$) recall fills a slot the user has just opened:
\begin{quote}\small
\textit{User}: ``There's a movie I've been looking forward to, and it's finally coming out, so I want to go see it.''\\
\textit{System}: ``Is it the one you mentioned before, like the new Predator, or something different this time?''
\end{quote}
A mistimed recall (appropriateness${=}0$) is factually grounded but disconnected from what the user just said:
\begin{quote}\small
\textit{User}: ``Thank you for saying that! See you next time! Good night.''\\
\textit{System}: ``Even though you're feeling exhausted from caring for your son's illness, your approach to taking small steps and focusing on self-care, like eating more vegetables, is really inspiring\ldots''
\end{quote}
Both reference real prior-session content; what differs is whether the user's immediately preceding turn invited the callback.

\section{Cost and Latency}
\label{app:cost}

Per-condition token usage and latency are summarized in main-text Table~\ref{tab:cost}. This appendix provides the methodology and the full latency-vs-satisfaction breakdown supporting the latency-confound check in \S\ref{sec:result1_null}.

\paragraph{Compute provenance.}
All experiments use commercial inference APIs (OpenAI \texttt{gpt-4.1-mini}, \texttt{gpt-5.4-nano}, and \texttt{gpt-5.5}); no local GPU training or fine-tuning was performed, and exact parameter counts for these closed-weight models are not publicly disclosed. Token-throughput cost (Table~\ref{tab:cost}) and wall-clock latency therefore serve as the computational-budget summary in lieu of GPU-hour figures.

\paragraph{Methodology.}
Cost numbers in Table~\ref{tab:cost} are computed using the OpenAI \texttt{gpt-4.1-mini} input-token rate over reconstructed per-session contexts; latency is wall-clock total per session in the deployed system. Token growth from month~1 to month~4 reflects the linear context-accumulation term that drives long-context cost: LC-100\% accumulates roughly $5.8\times$ more input tokens by month~4, while the importance-filtered RAG conditions stay flat or near-flat. Cost-per-memory-moment is computed as session cost divided by the observed 4\% moment rate, treating non-memory-moment sessions as cost without memory benefit.

\paragraph{Latency does not confound the capacity null.}
One alternative explanation for the satisfaction null is that low-latency conditions (Summary, RAG-$k\%$) might be rated higher \textit{because of} their shorter response times, masking a true capacity effect. The data do not support this. On the same 29-user / 1{,}270-session filtered subset used for satisfaction analysis, we computed per-session mean, maximum, and minimum total response latency and time-to-first-token (TTFT) across system turns and correlated each with within-user $z$-scored rating.
\textbf{Overall (Spearman):} mean total latency $\rho = -0.035$ ($p = .22$), max total latency $\rho = -0.012$ ($p = .67$), min total latency $\rho = -0.051$ ($p = .07$); mean TTFT $\rho = -0.026$ ($p = .36$), max TTFT $\rho = -0.001$ ($p = .96$), min TTFT $\rho = -0.051$ ($p = .07$).
\textbf{Within every condition:} $|\rho| \le 0.08$, all $p > .33$ (Summary-only $-0.04$; LC-10\% $-0.03$; LC-50\% $-0.03$; LC-100\% $-0.06$; RAG-10\% $+0.03$; RAG-50\% $-0.08$; RAG-100\% $+0.05$).
Notably, per-session \textit{maximum} latency---which captures the worst-case turn (single-turn maximum reaches $34.3$\,s in our deployment, $p_{95} = 6.5$\,s)---is the \textit{least} predictive of all six features: a ``one slow turn anchors the impression'' confound would predict max-turn latency to track satisfaction most strongly, whereas observed $\rho = -0.012$. The capacity null is not latency-mediated at the scale spanned by our deployment ($\sim$1.3\,s of mean-latency variance from Summary to LC-100\%, with single-turn maxes reaching tens of seconds).

\section{Data Release Notes}
\label{app:data_release}

We release (\url{https://huggingface.co/datasets/RuiSumida/memuse}; code at \url{https://github.com/ryuichi-sumida/memuse}): (i) the full deployment corpus (1{,}872 sessions across 40 users, 7 memory conditions, with per-session satisfaction ratings); (ii) the \textsc{MemUse} benchmark (72~de-identified instances with ground-truth facts and 316~questions, plus Natural Integration judgments and scoring prompts); and (iii) all evaluation and analysis code.
Consent procedures, redaction process, and privacy considerations are described in the Ethics Statement (main text).

\paragraph{License and intended use.}
The deployment corpus and \textsc{MemUse} benchmark are released under the \href{https://creativecommons.org/licenses/by-nc/4.0/}{Creative Commons Attribution-NonCommercial 4.0 International (CC BY-NC 4.0)} license; the evaluation and analysis code are released under the \href{https://opensource.org/licenses/MIT}{MIT} license. Both data artifacts are intended exclusively for non-commercial research use, consistent with the participant consent procedures (Ethics Statement); commercial redistribution of the conversational content is prohibited. Existing benchmarks compared against in this work (MemoryBank, LoCoMo, LongMemEval, LUFY, RealTalk) are used under their respective public-research licenses as released by their original authors. Commercial inference APIs (OpenAI's GPT family) are used in accordance with OpenAI's research-use terms of service.